\documentclass{article} 

\usepackage{amsmath,amssymb,bbm}
\usepackage{amsthm}
\usepackage{booktabs,multirow,graphicx,xcolor}
\usepackage{pifont}
\newcommand{\xmark}{\ding{55}}
\usepackage{colortbl}
\usepackage[ruled,vlined]{algorithm2e}

\usepackage{iclr2027_conference,times}

\usepackage{amsmath,amsfonts,bm}

\def\eqref#1{equation~\ref{#1}}

\def\1{\bm{1}}

\DeclareMathAlphabet{\mathsfit}{\encodingdefault}{\sfdefault}{m}{sl}
\SetMathAlphabet{\mathsfit}{bold}{\encodingdefault}{\sfdefault}{bx}{n}

\usepackage{hyperref}
\usepackage{url}
\usepackage{wrapfig}
\usepackage{subcaption}
\usepackage{float}
\usepackage{caption}

\newtheorem{proposition}{Proposition}

\newtheorem{corollary}{Corollary}

\title{\textsc{Cord}: Confidence-Anchored Order Residual Decoding for
Masked Diffusion Language Models}

\author{
  Chengwei Zhou\textsuperscript{1,*}\quad
  Zhaoyan Jia\textsuperscript{2,*}\quad
  Massoud Pedram\textsuperscript{2} \quad
  Gourav Datta\textsuperscript{1} \\
  \textsuperscript{1}Case Western Reserve University, Cleveland, USA \\
  \textsuperscript{2}University of Southern California, Los Angeles, USA \\
  {\small\textsuperscript{*}Equal contribution.}
}

\iclrfinalcopy 
\begin{document}

\maketitle

\begin{abstract}
The decoding order of masked diffusion language models (dLLMs) can substantially affect generation quality, yet common decoders typically apply the same reveal rule throughout generation. Confidence provides a strong default, but it measures how decisively a position can be predicted \emph{now}, not how useful revealing that position will be for subsequent decoding. We propose \textsc{Cord}, a lightweight prompt-conditioned policy that retains confidence decoding as a stable reference and learns when and how to depart from it through geometric residuals over generation direction,
proximity to revealed context, and two-sided infilling. The residuals adapt to both the prompt and the evolving partial response. Rather than sampling a new policy action at every decoding step, \textsc{Cord} samples a single
low-dimensional coefficient matrix per rollout, which induces an entire state-dependent reveal trajectory. This yields a contextual-bandit formulation trained directly from sequence-level rewards, without step-level credit
assignment or backbone updates. Across three dLLMs and five reasoning, code, and constraint-satisfaction benchmarks, \textsc{Cord} improves average task success from $32.0\%$ under confidence decoding to $48.2\%$, and further
improves over the best per-domain fixed residual. Behavioral audits show that the learned residuals induce substantial, task-dependent changes in reveal locality and two-sided infilling. \textsc{Cord} adds only $0.53$M trainable
parameters, requires no additional backbone forward passes, and incurs only $5.39\%$ inference latency overhead.
\end{abstract}

\vspace{-3mm}
\section{Introduction}\label{sec:intro}
\vspace{-2mm}

Masked diffusion language models (dLLMs) can reveal masked response positions
in different orders, unlike autoregressive models that generate strictly from
left to right. Each revealed token becomes part of the context for subsequent
predictions, so the reveal order can materially affect the final generation.
For example, \citet{kim2025trainworst} report that replacing random unmasking
with a top-two probability-margin rule increases Sudoku accuracy from below
$7\%$ to roughly $90\%$, while yielding much smaller gains on
natural-language reasoning. This disparity suggests that reveal order should
be treated as part of the decoding strategy, and that the appropriate strategy
can depend on the task. A common approach is to rank masked positions using a confidence statistic
computed from the current predictive distribution. Such scores are inexpensive,
since they are already available from the dLLM forward pass, and identify
positions whose current token prediction is most decisive. However, confidence
describes how predictable a position is \emph{now}; it does not measure how
useful revealing that position will be for later predictions. This distinction
depends on the geometry of the partially revealed response. Sequential
reasoning may benefit from extending an existing reasoning chain, whereas
structured generation may benefit from first revealing distant structural
anchors and then filling the gaps between them.

We verify this effect with frozen-backbone oracle experiments that control the
amount of revealed information. We reveal selected ground-truth tokens before
generation and compare them against the same number of randomly positioned
ground-truth tokens, after which the remaining response is completed using
standard confidence decoding. On GSM8K, revealing a correct causal prefix
increases accuracy from $37.2\%$ to as high as $67.4\%$, while token-matched
random anchors remain below $20\%$. On MBPP, revealing structural program
tokens outperforms matched random anchors by up to $14.4$ points. These results
show that, even at the same reveal budget, \emph{where} information is revealed
can substantially affect completion quality.

One way to exploit this structure is to learn the next reveal position
directly. Prior work uses reinforcement learning with sequence-level rewards
\citep{huang2025dcolt,zhao2025diffpo,jazbec2025unmask,du2025dultra}
or auxiliary planning networks \citep{peng2025p2,liu2024ddpd}. These
position-level formulations optimize a sequence of dependent discrete actions
from final-response feedback, resulting in a large action space and temporal
credit-assignment problem. Auxiliary planners can also add computation at
every decoding step, while jointly updating a multi-billion-parameter backbone
substantially increases training cost.

To adapt reveal order with substantially lower optimization and inference
overhead, we propose \textsc{Cord}. Instead of replacing confidence decoding
with a token-level planner, \textsc{Cord} keeps confidence as the base ranking
and learns bounded geometric corrections to it. \textsc{Cord} adjusts three
complementary properties of the reveal order: generation direction, proximity
to already revealed context, and preference for positions with revealed
context on both sides. A lightweight prompt-conditioned network predicts a
coefficient matrix, while features of the evolving partial response determine
how these corrections change during decoding. \textsc{Cord} can therefore
favor local continuation, non-local seeding, or two-sided infilling at
different stages of generation. When the residual coefficients are zero,
\textsc{Cord} reduces exactly to confidence-based (Top-2 margin) decoding, and the dLLM backbone
remains frozen.

\textsc{Cord} is trained by exploring in the low-dimensional coefficient space
rather than sampling a new reveal action at every step. For each prompt, the
policy samples one coefficient matrix that deterministically induces the
complete state-dependent reveal trajectory and receives a sequence-level
reward. This converts reveal-order learning into a contextual-bandit problem
and avoids step-level credit assignment. Across three masked diffusion models
and five benchmarks spanning mathematical reasoning, code generation, and
constraint satisfaction, \textsc{Cord} improves average task success from
$30.6\%$ under confidence-based decoding to $50.5\%$ and also outperforms the best
per-domain fixed geometric residual. Ablations show that prompt conditioning,
within-rollout adaptation, and all three geometric residuals contribute to the
gain. A behavioral analysis further shows task-dependent changes in reveal
locality and two-sided infilling. \textsc{Cord} requires no additional backbone
forward passes, adds only $0.53$M trainable parameters, and incurs $5.39\%$
inference latency overhead.

\vspace{-2mm}
\section{Related Work}\label{sec:related}
\vspace{-2mm}

\noindent\textbf{Masked diffusion language models.}
Absorbing-state discrete diffusion \citep{austin2021d3pm} has been simplified
into scalable masked-diffusion objectives
\citep{lou2024sedd,sahoo2024mdlm,shi2024md4,ou2024radd}, extended with variants
such as soft masking \citep{softmasked2026}, and scaled to instruction-following
models including LLaDA-8B-Instruct \citep{nie2025llada}, its
preference-optimized successor LLaDA-1.5 \citep{zhu2025llada15}, and the
autoregressively initialized Dream-7B \citep{ye2025dream}; see
\citet{survey2025dlm} for a survey. Unlike autoregressive models, masked
diffusion models admit multiple valid reveal orders at inference time. Although
this freedom is immaterial for an exactly fitted model, it can substantially
affect finite pretrained models. For example, \citet{kim2025trainworst} show
that a margin-based reveal rule raises Sudoku accuracy from below $7\%$ to
roughly $90\%$, while providing much smaller gains on natural-language
reasoning. This disparity motivates our view that reveal order should be treated
as a task- and state-dependent inductive bias rather than a fixed decoding
choice.

\vspace{-3mm}
\paragraph{Reveal-order strategies for MDMs.}
Existing approaches range from hand-designed reveal rules to learned planners. Confidence-ranked decoding descends from MaskGIT \citep{chang2022maskgit}; modern dLLMs use max probability \citep{ye2025dream}, top-2 margin \citep{kim2025trainworst}, negative entropy, or block-wise schedules, often in combination with caching, entropy-bounded thresholds, or adaptive block selection \citep{wu2025fastdllm,benhamu2025entropy,israel2025apd,dsb2026}.
Recent work also identifies limitations of purely confidence-driven ordering, including boundary and trivial-token biases \citep{huang2026biases}, premature deferral of high-entropy connectives that can alter reasoning trajectories \citep{ni2026flexibility}, and reduced sample diversity caused by coupling token prediction and reveal order \citep{olausson2026temperatures}. These findings motivate augmenting confidence with information about the geometry of the partially revealed response rather
than replacing confidence entirely. Learned methods instead optimize reveal order more directly. Some introduce
explicit planners for unmasking or remasking
\citep{peng2025p2,liu2024ddpd,sahoo2025papl,zhou2026attnsampler, liu2026nisampling,adaptiveorder2026}, while others distill oracle orders from ground-truth margins \citep{asano2026where} or optimize reveal decisions using terminal or pathwise rewards \citep{huang2025dcolt,zhao2025diffpo,jazbec2025unmask,du2025dultra,
sas2026scheduling}. These approaches typically operate at the level of
positions or decoding steps and therefore face a large action space and
long-horizon credit assignment; many also modify or jointly train the backbone. CORD takes a different route. We keep the pretrained confidence-based decoder as a fixed reference and learn a low-dimensional, prompt-conditioned residual over three interpretable geometric axes: direction, proximity to revealed context, and two-sided infilling. The resulting $24$-dimensional coefficient program controls how these residuals evolve with the decoding state, while the dLLM backbone remains frozen. Exploration occurs once per rollout in coefficient space, after which the complete reveal trajectory is deterministic, yielding a contextual-bandit formulation rather than a step-level planning problem.

\begin{figure}[t]
\centering
\includegraphics[trim={80 170 82 155}, clip, width=\linewidth]{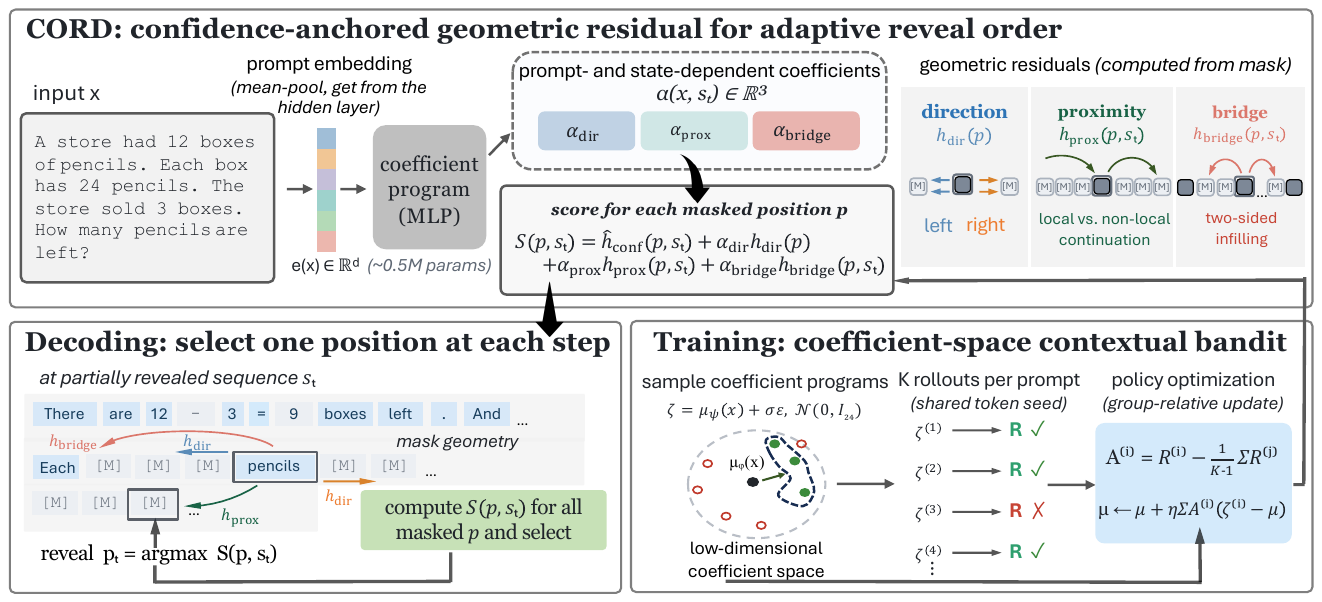}
\vspace{-3mm}
\caption{
\textbf{\textsc{Cord}.}
\emph{Top:} a prompt-conditioned MLP produces state-dependent geometric
residuals: direction, proximity, and bridge, that modify confidence-based
reveal scores while keeping the dLLM frozen.
\emph{Bottom left:} CORD reveals the highest-scoring masked position at each
step.
\emph{Bottom right:} training samples one low-dimensional coefficient program
per rollout and optimizes it from sequence-level rewards as a contextual bandit.
}
\label{fig:main}
\vspace{-6mm}
\end{figure}

\section{Method}
\label{sec:method}
\vspace{-2mm}
\subsection{From Task-Specific Ordering Biases to Dynamic Decoding}
\label{sec:motivation}
\vspace{-2mm}

\begin{wrapfigure}{r}{0.48\linewidth}
    \centering
    \vspace{-4mm}
    \includegraphics[trim={120 190 160 190}, clip, width=\linewidth]{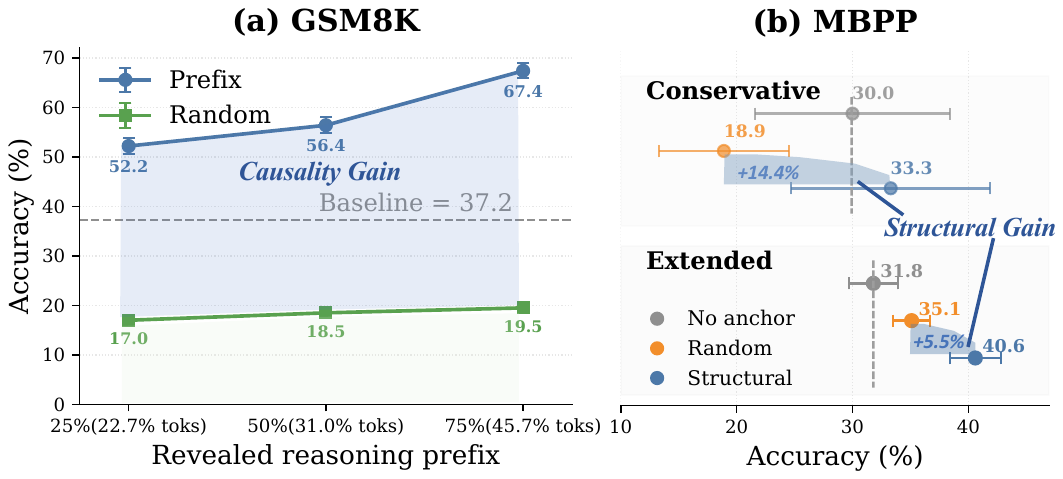}
    \vspace{-6mm}
    \caption{
    Oracle reveal diagnostics with frozen LLaDA. GSM8K favors a contiguous
    reasoning prefix, while MBPP benefits from structural anchors.
    \emph{Conservative} uses core control-flow tokens (e.g. \texttt{if}, \texttt{else}, \texttt{for}); \emph{Extended} additionally
    includes function (\texttt{def}, \texttt{class}), delimiter, and assignment tokens.
    }
    \label{fig:oracle_diagnostic}
    \vspace{-6mm}
\end{wrapfigure}

Given a prompt $x$, a masked diffusion language model (dLLM) with frozen parameters $\theta$ predicts token distributions over all masked response positions and must decide \emph{which position to reveal next}. We leave token prediction unchanged and study only this reveal-order decision.

We first ask whether different tasks actually prefer different forms of partial conditioning. To isolate this effect from learning, we conduct two \emph{oracle} diagnostics with the dLLM backbone frozen and reveal selected ground-truth tokens before generation. On GSM8K, we reveal the first $25\%$, $50\%$, or $75\%$ of the safe reasoning \emph{lines}, while excluding the final reasoning line and answer,
and compare against revealing the same number of randomly located ground-truth tokens. On MBPP, we instead reveal the structural skeleton of the canonical program, e.g., \texttt{if}, \texttt{else}, \texttt{for}, \texttt{while}, and \texttt{return}, and compare against a token-budget-matched random control. In both cases, the remaining positions are completed by standard confidence LLaDA decoding. 
Figure~\ref{fig:oracle_diagnostic} shows clear task-dependent preferences. On GSM8K, revealing a causal prefix raises accuracy from no-oracle $37.2\%$ to as high as $67.4\%$, while matched random anchors remain below $20\%$. On MBPP, structural anchors outperform matched random anchors by up to $14.4$ points. These results suggest that reasoning favors coherent prefix expansion, whereas code generation benefits from sparse structural scaffolds.

This suggests that no single reveal bias is uniformly appropriate. Sequential
reasoning may favor coherent local continuation, while structured generation
may benefit from non-local anchors that support later infilling. Moreover, the
preferred bias can change during generation as newly revealed tokens reshape
both the confidence-based landscape and the geometry of the remaining masked region.

CORD therefore retains confidence-based decoding as a stable default and
learns only a bounded \emph{geometric residual} that determines when and how
to deviate from it. The residual can adjust generation direction, trade off
local continuation against non-local seeding, and favor two-sided infilling,
while depending on both the prompt and the evolving decoding state.

\subsection{Confidence-Anchored Reveal Geometry}
\label{sec:heuristics}

Let $s_t$ denote the decoding state at step $t$, $\mathcal{M}(s_t)$ the set of still-masked response positions, and $R_{\mathrm{resp}}(s_t)$ the set of already revealed response positions. For each $p\in\mathcal{M}(s_t)$, the frozen dLLM predicts
\begin{equation}
\pi_p(v\mid s_t)
=
p_\theta(y_p=v\mid s_t).
\label{eq:token_dist}
\vspace{-2mm}
\end{equation}

CORD retains confidence-based decoding as its base policy and learns only geometric residuals that modify the reveal ranking. Let $v_1(p)$ and $v_2(p)$ denote the two highest-probability non-special tokens at $p$, and let $\operatorname{idx}(p)\in\{1,\ldots,L\}$ denote its response position. We define
\begin{equation}
\begin{aligned}
h_{\rm anchor}(p,s_t)
&=
\pi_p(v_1(p)\mid s_t)
-
\pi_p(v_2(p)\mid s_t),
\\
h_{\mathrm{dir}}(p)
&=
2\frac{\operatorname{idx}(p)-1}{L-1}-1,
\\
h_{\mathrm{prox}}(p,s_t)
&=
\begin{cases}
\displaystyle
\min_{r\in R_{\mathrm{resp}}(s_t)}
\left|
\operatorname{idx}(p)-\operatorname{idx}(r)
\right|,
&
R_{\mathrm{resp}}(s_t)\neq\varnothing,
\\
0,
&
R_{\mathrm{resp}}(s_t)=\varnothing,
\end{cases}
\\
h_{\mathrm{bridge}}(p,s_t)
&=
\mathbf{1}
\left[
\exists\, r_\ell,r_r\in R_{\mathrm{resp}}(s_t):
\operatorname{idx}(r_\ell)
<
\operatorname{idx}(p)
<
\operatorname{idx}(r_r)
\right].
\end{aligned}
\label{eq:reveal_biases}
\vspace{-2mm}
\end{equation}

These signals have complementary interpretations. $h_{\rm anchor}$ recovers the conventional confidence-based (Top-2 margin) decoder. For $h_{\mathrm{dir}}$, a negative coefficient favors earlier positions
(left-to-right), whereas a positive coefficient favors later positions (right-to-left).
For $h_{\mathrm{prox}}$, a negative coefficient favors positions near already revealed response tokens and therefore encourages local continuation, whereas a positive coefficient favors distant positions and promotes non-local seeding. Finally, a positive coefficient on $h_{\mathrm{bridge}}$ favors positions with
revealed response context on both sides and therefore encourages infilling. Prompt tokens are excluded from $R_{\mathrm{resp}}$ so that the prompt--response boundary does not introduce an artificial positional bias.

Because these signals have different numerical ranges, we normalize each one over the currently masked positions:
$
\widetilde h_m(p,s_t)
=
\begin{cases}
\dfrac{
h_m(p,s_t)-\mu_m^{(t)}
}{
\sigma_m^{(t)}
},
&
\sigma_m^{(t)}\geq\epsilon_{\mathrm{var}},
\\[8pt]
0,
&
\text{otherwise},
\end{cases}
\quad
\epsilon_{\mathrm{var}}=10^{-4},
\label{eq:residual_norm}
$
where $\mu_m^{(t)}$ and $\sigma_m^{(t)}$ are computed over $p\in\mathcal{M}(s_t)$. Near-constant signals are therefore suppressed rather than amplified by normalization.

\subsection{Dynamic Geometric Residuals}
\label{sec:dynamic_mixture}

CORD augments the normalized anchor score with a learned geometric residual:
\begin{equation}
S(p,s)
=
\widetilde h_{\rm anchor}(p,s)
+
\alpha_{\mathrm{dir}}(x,s)
\widetilde h_{\mathrm{dir}}(p)
+
\alpha_{\mathrm{prox}}(x,s)
\widetilde h_{\mathrm{prox}}(p,s)
+
\alpha_{\mathrm{bridge}}(x,s)
\widetilde h_{\mathrm{bridge}}(p,s).
\label{eq:residual_score}
\end{equation}
At each decoding step, CORD reveals the single masked position with the largest score:
$
p^\star
=
\arg\max_{p\in\mathcal{M}(s)}
S(p,s).
\label{eq:reveal_position}
$
The anchor term remains active at every step, while the residuals adjust the reveal geometry. In particular, $\alpha_{\mathrm{dir}}$ controls left-to-right versus right-to-left preference,
$\alpha_{\mathrm{prox}}$ trades off local continuation against non-local seeding, and $\alpha_{\mathrm{bridge}}$ controls the preference for positions with revealed context on both sides. Unlike hard routing, these residuals only modify the ranking and never exclude positions from consideration.

To let these preferences evolve during decoding without introducing a new stochastic policy action at every step, CORD predicts a single coefficient program. Let $B=3$ denote the residual channels
$\{\mathrm{dir},\mathrm{prox},\mathrm{bridge}\}$ and let $J$ denote the number of state-basis functions. For a prompt-conditioned matrix
$
C(x)\in\mathbb{R}^{B\times J},
$
the residual coefficients are
\begin{equation}
\boldsymbol{\alpha}(x,s)
=
\boldsymbol{\alpha}_{\max}
\odot
\tanh
\left(
C(x)\mathbf{f}(s)
\right),
\label{eq:residual_coefficients}
\end{equation}
where $\boldsymbol{\alpha}_{\max}\in\mathbb{R}_{+}^{B}$ bounds the magnitude of each correction. We use the state basis
\vspace{-2mm}
\begin{equation}
\mathbf{f}(s)
=
\left[
\underbrace{
1,\;
t,\;
\sin(\pi t),\;}_{\text{defusion step bias}}
\underbrace{
F_{\mathrm{mean}}(s),\;
F_{\mathrm{conc}}(s),\;
F_{\mathrm{comp}}(s),\;
F_{\mathrm{bound}}(s),\;
F_{\mathrm{gap}}(s)}_{\text{state dependencies}}
\right]^{\top},
\label{eq:basis}
\vspace{-2mm}
\end{equation}
where
$
t
=
\frac{|R_{\mathrm{resp}}(s)|}{L}
$
is the fraction of the response already revealed. The first two state features summarize the confidence-based landscape:
\vspace{-1mm}
\begin{equation}
F_{\mathrm{mean}}(s)
=
\frac{1}{|\mathcal{M}(s)|}
\sum_{p\in\mathcal{M}(s)}
h_{\rm anchor}(p,s),
\quad
F_{\mathrm{conc}}(s)
=
Q_{0.9}\!\left(h_{\rm anchor}(\cdot,s)\right)
-
Q_{0.5}\!\left(h_{\rm anchor}(\cdot,s)\right).
\label{eq:confidence_features}
\vspace{-1mm}
\end{equation}
They measure the overall confidence-based level and whether a small subset of masked positions is substantially easier than the rest. The remaining features summarize the reveal geometry:
\begin{equation}
F_{\mathrm{comp}}(s)
=
\frac{N_{\mathrm{comp}}(s)}{L},
\qquad
F_{\mathrm{bound}}(s)
=
\frac{|\mathcal{B}(s)|}{|\mathcal{M}(s)|},
\qquad
F_{\mathrm{gap}}(s)
=
\frac{G_{\max}(s)}{L},
\label{eq:geometry_features}
\end{equation}
where $N_{\mathrm{comp}}(s)$ is the number of contiguous revealed-response components, $\mathcal{B}(s)$ is the set of masked positions with revealed context on both sides, and $G_{\max}(s)$ is the length of the largest contiguous masked run.

A single $B\times J$ matrix therefore determines how the three geometric residuals evolve over the entire rollout. Once $C(x)$ is fixed, the residual coefficients are deterministic functions of the current decoding state. The coefficient predictor is zero-initialized, so CORD begins exactly from confidence-based decoding and learns only residual departures from it.

\subsection{Prompt-Conditioned Coefficient Policy}
\label{sec:coefficient_policy}

The geometric residuals should depend on the input, since different prompts may favor different reveal-order corrections even within the same task. CORD therefore predicts the coefficient program directly from the prompt. The first backbone forward pass already processes the prompt together with the masked response block, so we mean-pool the intermediate prompt hidden states to obtain a representation $e(x)\in\mathbb{R}^{d}$ without requiring an additional backbone forward pass.

A lightweight policy network predicts the mean coefficient vector:
\vspace{-1mm}
\begin{equation}
\boldsymbol{\mu}_{\psi}:\;
e(x)
\xrightarrow{
f_{\psi}:\;
\mathrm{Linear}(d\!\to\!128)
\to
\mathrm{SiLU}
\to
\mathrm{Linear}(128\!\to\!n)
}
\boldsymbol{\mu}(x)
\in\mathbb{R}^{n},
\qquad
n=BJ.
\label{eq:policy_mean}
\vspace{-2mm}
\end{equation}

We use a two-layer MLP with a $128$-dimensional hidden layer and SiLU activation. The output $\boldsymbol{\mu}_{\psi}(x)$ is reshaped into a coefficient matrix $C(x)\in\mathbb{R}^{B\times J}$. In our configuration, $B=3$ residual channels and $J=8$ state-basis functions, giving $n=24$.
The final linear layer is initialized to zero, so CORD initially reduces exactly to confidence-based decoding. This parameterization separates prompt-level and state-level adaptation. The prompt determines the coefficient program $C(x)$, while the evolving state basis $\mathbf{f}(s)$ determines how this program is expressed as geometric residuals during decoding. Thus, a single prompt-conditioned coefficient matrix specifies an entire state-dependent reveal strategy without requiring a new policy decision at every decoding step. During training, CORD explores around the predicted mean in coefficient space; at inference, it directly uses the mean program $C(x)$.

\subsection{Training CORD in Coefficient Space}
\label{sec:bandit}
\vspace{-2mm}
There is no ground-truth reveal order with which to supervise CORD directly: the quality of an ordering strategy is only known after the resulting response has been generated and evaluated. We therefore train the coefficient policy from sequence-level rewards. A token-level policy would make a stochastic decision at every decoding step, creating a long sequence of dependent actions and requiring temporal credit assignment. Instead, CORD places exploration in the low-dimensional coefficient space that parameterizes the entire dynamic reveal strategy.

For each prompt $x$, the policy predicts a mean coefficient vector $\boldsymbol{\mu}(x)\in\mathbb{R}^{n}$, where $n=BJ$. During training, we sample a perturbed coefficient vector
\vspace{-2mm}
\begin{equation}
C_{\boldsymbol{\zeta}}
=
\operatorname{reshape}_{B\times J}
\left(
\boldsymbol{\zeta}
\right),
\qquad
\boldsymbol{\zeta}
=
\boldsymbol{\mu}(x)
+
\sigma\boldsymbol{\epsilon},
\qquad
\boldsymbol{\epsilon}
\sim
\mathcal{N}\left(\mathbf{0},I_n\right).
\label{eq:coefficient_sample}
\vspace{-2mm}
\end{equation}

The sampled matrix $C_{\boldsymbol{\zeta}}$ is held fixed for the entire rollout. At each decoding state $s$, it is combined with the current state basis $\mathbf{f}(s)$ through Eq.~\ref{eq:residual_coefficients} to produce the geometric residual coefficients
$\boldsymbol{\alpha}(x,s)$. Although these coefficients evolve with the decoding state, once $\boldsymbol{\zeta}$ is sampled, all subsequent reveal decisions are deterministic.

Each prompt therefore defines a context $x$, the policy takes a single stochastic action $\boldsymbol{\zeta}$, and the resulting decoding trajectory receives one sequence-level reward $r$:
\vspace{-1mm}
\begin{equation}
x
\;\longrightarrow\;
\boldsymbol{\zeta}
\;\longrightarrow\;
\text{full decoding rollout}
\;\longrightarrow\;
r.
\label{eq:bandit_view}
\vspace{-2mm}
\end{equation}

CORD can thus be trained as a contextual bandit rather than as a long-horizon sequential policy. In our configuration, the stochastic action space has only $BJ=3\times8=24$ dimensions, while still inducing a state-dependent reveal strategy over the entire rollout. This avoids step-level credit assignment and keeps the frozen dLLM outside the policy-gradient computation.

At inference time, exploration is removed and we use the predicted mean coefficient program
$
C(x)
=
\operatorname{reshape}_{B\times J}
\left(
\boldsymbol{\mu}(x)
\right).
\label{eq:inference_coeff}
$
Decoding is then deterministic using
Eqs.~\ref{eq:residual_coefficients} and~\ref{eq:residual_score}.

\subsection{Sequence-Level Reward}
\label{sec:reward}
\vspace{-2mm}
CORD is trained from sequence-level feedback rather than supervision on individual reveal decisions. For rollout $i$, we use
\begin{equation}
r^{(i)}
=
\mathbf{1}
\left[
\text{answer}^{(i)}
\text{ is correct}
\right]
+
\lambda_{\mathrm{fmt}}
\mathbf{1}
\left[
\text{output}^{(i)}
\text{ is parseable}
\right]
+
\lambda_{\mathrm{ans}}
\overline{\log p}_{\mathrm{ans}}^{(i)}.
\label{eq:reward}
\vspace{-3mm}
\end{equation}

The correctness term provides the primary supervision and directly evaluates whether the induced reveal strategy produces a successful final response. We do not prescribe a target reveal trajectory; instead, the coefficient policy is optimized according to the quality of the complete generation.

We use a small formatting reward, with $\lambda_{\mathrm{fmt}}=0.1$, to distinguish malformed outputs from valid but incorrect responses. When additional graded feedback is useful, the optional
answer-probability term re-masks the reference answer span and measures the mean log-probability assigned to the reference answer tokens. This requires one additional backbone forward pass.
Task-specific verifiers define correctness. Mathematical responses are evaluated by answer extraction and symbolic equivalence, generated code is executed against the corresponding test cases in a resource-limited sandbox, and Sudoku outputs are checked for clue preservation together with row, column, and box constraints. Training and evaluation use the same verifier implementation.

\subsection{Group-Relative Policy Gradient in Coefficient Space}
\label{sec:group_pg}
\vspace{-2mm}
For each prompt $x$, we generate a group of $K$ sibling rollouts by sampling
$
\boldsymbol{\zeta}^{(i)}
\sim
q_{\psi}
\left(
\boldsymbol{\zeta}\mid x
\right)
=
\mathcal{N}
\left(
\boldsymbol{\mu}_{\psi}(x),
\sigma^2 I_n
\right),
i=1,\ldots,K.
\label{eq:policy_dist}
$
Each sampled vector is reshaped into a coefficient program $C_{\boldsymbol{\zeta}^{(i)}}$ and held fixed throughout the corresponding rollout.
The log-density of a sampled coefficient vector is
\vspace{-2mm}
\begin{equation}
\log
q_{\psi}
\left(
\boldsymbol{\zeta}^{(i)}
\mid x
\right)
=
-
\frac{
\left\|
\boldsymbol{\zeta}^{(i)}
-
\boldsymbol{\mu}_{\psi}(x)
\right\|_2^2
}{
2\sigma^2
}
-
\frac{n}{2}
\log
\left(
2\pi\sigma^2
\right),
\label{eq:gaussian_logprob}
\end{equation}
with score-function gradient
\vspace{-2mm}
\begin{equation}
\vspace{-2mm}
\nabla_{\psi}
\log
q_{\psi}
\left(
\boldsymbol{\zeta}^{(i)}
\mid x
\right)
=
\left(
\frac{
\boldsymbol{\zeta}^{(i)}
-
\boldsymbol{\mu}_{\psi}(x)
}{
\sigma^2
}
\right)^{\top}
\nabla_{\psi}
\boldsymbol{\mu}_{\psi}(x).
\label{eq:score_gradient}
\vspace{-0.5mm}
\end{equation}
The gradient therefore never passes through the discrete reveal trajectory: CORD learns which regions of coefficient space induce successful decoding strategies rather than differentiating through the reveal decisions themselves.
To reduce prompt-dependent reward variation without introducing a learned critic, we compare sibling rollouts generated from the same prompt. For rollout $i$, we use the leave-one-out group-relative advantage
\vspace{-2mm}
\begin{equation}
A^{(i)}
=
r^{(i)}
-
\frac{1}{K-1}
\sum_{\substack{j=1 \\ j\neq i}}^{K}
r^{(j)}.
\label{eq:group_advantage}
\vspace{-2mm}
\end{equation}
The resulting advantage measures whether a coefficient perturbation performs better or worse than alternative perturbations for the same prompt.
We do not normalize advantages by their within-group standard deviation. For binary or nearly binary verifier rewards, such normalization can disproportionately amplify groups containing only one success or one failure. If all sibling rollouts receive the same reward, then
$
A^{(i)}=0,
\forall i\in\{1,\ldots,K\},
$
and the group contributes no policy-gradient signal. The resulting objective is
$\mathcal{L}_{\mathrm{PG}}
=
-
\frac{1}{K}
\sum_{i=1}^{K}
A^{(i)}
\log
q_{\psi}
\left(
\boldsymbol{\zeta}^{(i)}
\mid x
\right).$
The corresponding update on the coefficient mean has the intuitive form
$
\boldsymbol{\mu}
\leftarrow
\boldsymbol{\mu}
+
\frac{\eta}{K\sigma^2}
\sum_{i=1}^{K}
A^{(i)}
\left(
\boldsymbol{\zeta}^{(i)}
-
\boldsymbol{\mu}
\right),
\label{eq:intuitive_update}
$
so successful perturbations pull the predicted coefficient program toward themselves, while unsuccessful perturbations push it away.
Finally, sibling rollouts for the same prompt share all randomness other than the sampled coefficient vector. When token generation is stochastic, we use common random numbers across siblings; when token generation is greedy, token selection is deterministic. This reduces variance in within-prompt comparisons by making reward differences primarily reflect the sampled reveal-order program. We perform one on-policy update per rollout batch and require neither a learned critic nor importance-ratio correction.

\vspace{-2mm}
\section{Experiments}
\label{sec:exp}
\vspace{-2mm}

\subsection{Experimental Setup}
\label{sec:setup}
\vspace{-2mm}
We evaluate CORD on three frozen masked diffusion language models: LLaDA-8B-Instruct~\citep{nie2025llada}, LLaDA-1.5~\citep{zhu2025llada15}, and Dream-7B~\citep{ye2025dream}. Unless stated otherwise, we train a separate lightweight CORD policy for each backbone while keeping all dLLM parameters frozen. Evaluating the same ordering framework across three independently trained checkpoints tests whether CORD exploits a general property of masked diffusion decoding rather than a backbone-specific artifact.
We evaluate three qualitatively different forms of dependency structure:
mathematical reasoning on GSM8K~\citep{cobbe2021gsm8k} and MATH~\cite{hendrycksmath2021}, code generation on HumanEval~\citep{chen2021evaluating} and MBPP~\cite{austin2021program},
and constraint satisfaction on Sudoku~\citep{ye2025autoregressiondiscretediffusioncomplex}. These domains are deliberately
different in the dependencies they impose on the response: reasoning is largely
sequential, code combines global structure with local completion, and Sudoku
contains dense non-local constraints. This diversity allows us to test the
central hypothesis of CORD, that no single reveal bias is uniformly optimal
across tasks. Dataset construction, splits, prompt formatting, and verifier
details are provided in Appendix~\ref{app:setup}.


\vspace{-2mm}
\subsection{Main Results: CORD Generalizes Across MDMs}
\label{sec:main_results}
\vspace{-2mm}
Table~\ref{tab:main} compares CORD with representative reveal-order baselines
across three masked diffusion 
language models. We include conventional
confidence-, entropy-, and margin-decoding, stronger existing decoding
methods, and a fixed-residual control that augments confidence-based with the same
geometric signals used by CORD but does not adapt them across prompts or
decoding states.
\begin{table*}[t]
\centering
\caption{
\textbf{Main results across masked diffusion language models.}
Task success (\%, higher is better). HumanEval and MBPP report pass@1;
all other tasks report accuracy. The fixed residual uses the same geometric
signals as CORD but keeps their coefficients constant throughout decoding.
}
\vspace{-2mm}
\label{tab:main}
\fontsize{7pt}{6pt}\selectfont
\setlength{\tabcolsep}{5pt}
\renewcommand{\arraystretch}{1.08}

\begin{tabular}{llcccccc}
\toprule
Backbone
& Decoding policy
& GSM8K
& MATH500
& HumanEval
& MBPP
& Sudoku
& Avg. \\
\midrule

\multirow{10}{*}{LLaDA-8B-Instruct}
& Confidence
& 58.7 & 18.0 & 27.8 & 38.4 & 17.1 & 32.0 \\
& Entropy
& 59.4 & 12.5 & 27.5 & 41.7 & 3.3 & 28.9 \\
& Margin~\citep{kim2025trainworst}
& 58.4 & 22.2 & 28.8 & 42.1 & 25.8 & 35.5 \\
& EB-Sampler~\citep{benhamu2025entropy}
& 61.4 & 12.7 & 27.1 & 42.7 & 33.8 & 35.5 \\
& Semi-AR$^\dagger$~\citep{nie2025llada}
& 76.8 & 28.1 & 38.9 & 45.2 & 0.3 & 37.9 \\
& UNCODE~\citep{huang2026biases}
& \textbf{79.2} & 34.8 & 42.1 & 47.8 & 29.8 & 46.7 \\
\cmidrule(lr){2-8}
\rowcolor{gray!20}
& Best per-domain fixed residuals (\textbf{Ours})
& 78.4 & 34.9 & 41.9 & 48.9 & 35.7 & 48.0 \\
\rowcolor{gray!20}
& \textbf{CORD}
& 78.2 & \textbf{34.9} & \textbf{42.7} & \textbf{48.9} & \textbf{36.1} & \textbf{48.2} \\
\midrule

\multirow{10}{*}{LLaDA-1.5-8B}
& Confidence
& 60.7 & 19.5 & 25.1 & 42.1 & 19.1 & 33.3 \\
& Entropy
& 60.4 & 12.3 & 28.1 & 42.7 & 0.4 & 28.8 \\
& Margin
& 58.7 & 22.9 & 26.7 & 43.1 & 31.2 & 36.5 \\
& EB-Sampler
& 61.0 & 16.6 & 32.7 & 42.7 & 0.4 & 30.7 \\
& Semi-AR$^\dagger$
& 79.1 & 33.8 & 37.7 & 45.9 & 0.4 & 39.4 \\
& UNCODE
& 82.2 & 37.4 & 46.3 & 49.9 & 33.4 & 49.8 \\
\cmidrule(lr){2-8}
\rowcolor{gray!20}
& Best per-domain fixed residuals (\textbf{Ours})
& 81.8 & \textbf{37.6} & 46.8 & 46.8 & 36.1 & 49.8 \\
\rowcolor{gray!20}
& \textbf{CORD}
& \textbf{82.6} & 37.4 & \textbf{46.9}
& \textbf{50.3} & \textbf{36.2} & \textbf{50.7} \\
\midrule

\multirow{9}{*}{Dream-7B}
& Confidence
& 39.5 & 10.9 & 29.1 & 31.6 & 20.7 & 26.4 \\
& Entropy
& 68.7 & 21.7 & 57.9 & 47.3 & 33.6 & 45.8 \\
& Margin
& 38.4 & 17.1 & 29.5 & 35.2 & 26.9 & 29.4 \\
& EB-Sampler
& 58.8 & 17.9 & 37.6 & 44.1 & 0.5 & 31.8 \\
& Semi-AR$^\dagger$
& 72.1 & 25.7 & 56.4 & 48.1 & 0.0 & 40.5 \\
\cmidrule(lr){2-8}
\rowcolor{gray!20}
& Best per-domain fixed residuals (\textbf{Ours})
& 80.2 & 35.7 & \textbf{58.6} & 49.7 & 37.0 & 52.2 \\
\rowcolor{gray!20}
& \textbf{CORD}
& \textbf{80.2} & \textbf{36.1} & 58.4
& \textbf{50.7} & \textbf{37.1} & \textbf{52.5} \\
\bottomrule
\end{tabular}

\vspace{1mm}
\raggedright
{\scriptsize
$^\dagger$ Semi-autoregressive decoding.
}
\vspace{-5mm}
\end{table*}
Across the three backbones, CORD improves average task success from
$30.6\%$ under confidence decoding to $50.5\%$, a gain of $19.9$ points.
It also remains competitive with or outperforms stronger reveal-order baselines
across reasoning, code generation, and Sudoku.
More importantly, CORD improves over the best per-domain fixed residuals from
$50.0\%$ to $50.5\%$ on average. Because this control uses the same direction,
proximity, and bridge signals, the remaining gain isolates the benefit of
adapting the residuals to the prompt and evolving decoding state. The consistent
performance across LLaDA-8B-Instruct, LLaDA-1.5-8B, and Dream-7B further
suggests that this adaptation transfers across differently trained MDMs.

\vspace{-2mm}
\subsection{Ablation}
\vspace{-2mm}
\noindent\textbf{Does Reveal Geometry Need to Adapt?} We evaluate adaptation variants on all five benchmarks and all three backbones,
retraining the coefficient policy for each variant under the same optimization
setup. The fixed-residual variant uses one shared residual across all prompts
and decoding steps; prompt only predicts an input-specific but static residual;
\begin{wraptable}{r}{0.47\textwidth}
\vspace{-2mm}
\centering
\captionof{table}{
\textbf{Decomposing adaptation in CORD.}
Mean task success (\%) across all benchmarks.
}
\vspace{-3mm}
\label{tab:adaptation}

\fontsize{7pt}{5.5pt}\selectfont
\setlength{\tabcolsep}{1pt}
\renewcommand{\arraystretch}{1.05}
\begin{tabular}{lcccc}
\toprule
Variant
& LLaDA-Inst.
& LLaDA-1.5
& Dream
& Avg. \\
\midrule
Confidence only
& 32.0 & 33.3 & 26.4 & 30.6 \\
\midrule
No prompt (fixed residual)
& 48.0 & 49.8 & 40.5 & 46.1 \\
Prompt only (static)
& 42.8 & 48.1 & 47.1 & 46.0 \\
Time only
& 46.3 & 45.1 & 49.2 & 46.9 \\
State only
& 47.1 & 46.4 & 33.1 & 42.2 \\
\rowcolor{gray!20}
\textbf{Full CORD}
& \textbf{48.2} & \textbf{50.7}
& \textbf{52.5} & \textbf{50.5} \\
\bottomrule
\end{tabular}
\vspace{-4mm}
\end{wraptable}
time only retains the explicit decoding-progress basis but removes state
features; and state only keeps state features while removing the temporal basis. As shown in Table~\ref{tab:adaptation}, full CORD achieves the best average
performance, improving over the fixed residual by $4.4\%$ and over the
prompt-only variant by $4.5\%$. Neither time only ($46.9\%$) nor state only
($42.2\%$) matches the full model ($50.5\%$), indicating that effective reveal
geometry benefits from both decoding progress and feedback from the evolving
partial response. The gains across all three backbones further suggest that this
adaptation is not specific to a single MDM.

\noindent\textbf{Are the Geometric Residuals Complementary?} We ablate the three residual channels on LLaDA-8B-Instruct using one
representative task from each domain: GSM8K for reasoning, MBPP for code, and
Sudoku for constraint satisfaction. For each variant, we remove one residual
channel and retrain the coefficient policy under the same training setup,
leaving all other components unchanged; 
\begin{wraptable}{r}{0.35\textwidth}
\centering
\vspace{-2mm}
\captionof{table}{
\textbf{Ablation of geometric residuals.}
Task success (\%) on LLaDA-8B-Instruct.
}
\vspace{-2mm}
\label{tab:residual_ablation}

\fontsize{7pt}{6pt}\selectfont
\setlength{\tabcolsep}{2pt}
\renewcommand{\arraystretch}{1.05}

\begin{tabular}{lcccc}
\toprule
Variant
& GSM8K
& MBPP
& Sudoku
& Avg. \\
\midrule
$-\mathrm{Direction}$
& 57.2 & 43.7 & 35.3 & 45.4 \\
$-\mathrm{Proximity}$
& 70.2 & 44.8 & 31.8 & 48.9 \\
$-\mathrm{Bridge}$
& 72.4 & 41.4 & 16.6 & 43.5 \\
\rowcolor{gray!20}
\textbf{Full CORD}
& \textbf{78.2} & \textbf{48.9}
& \textbf{36.1} & \textbf{54.4} \\
\bottomrule
\end{tabular}
\end{wraptable}
the reported average is over these three tasks. As shown in Table~\ref{tab:residual_ablation}, removing any channel reduces average performance, with drops of $9.0\%$, $5.5\%$, and $10.9\%$ for
direction, proximity, and bridge, respectively. Their effects are also
task dependent: direction is especially important on GSM8K, while bridge is
critical on Sudoku. Proximity provides a smaller but consistent contribution
across all three tasks. These results support using complementary geometric
corrections rather than a single alternative reveal bias.

\vspace{-2mm}
\subsection{What Reveal Geometry Does CORD Learn?}
\label{sec:behavior}
\vspace{-2mm}
We conduct a post-hoc behavioral audit of the trained CORD policy on GSM8K,
MBPP, and Sudoku, without additional training or oracle information based on LLaDA-8B-Instruct. For each
prompt, we run CORD and an independent confidence-decoding baseline under the
same generation configuration: $256$ generation steps, response length $256$,
greedy token decoding ($T=0$), and EOS suppression. The backbone remains frozen
throughout. 

We measure three aspects of the realized reveal trajectory. First, the
\emph{confidence-override rate} records how often CORD selects a different
position from confidence alone. Importantly, both decisions are computed on the
same CORD state $s_t$, avoiding confounding from trajectories that have already
diverged. Second, reveal locality is quantified by the normalized jump distance
between consecutive reveals, reported as
$\Delta J=J_{\mathrm{CORD}}-J_{\mathrm{conf}}$; negative values indicate more
local decoding and positive values more non-local decoding. Third,
\emph{bridge enrichment} measures whether CORD selects positions with revealed
context on both sides more often than their availability among the remaining
masked positions would predict. 
\begin{wrapfigure}{r}{0.47\linewidth}
\vspace{-3mm}
    \centering
    \includegraphics[width=\linewidth]{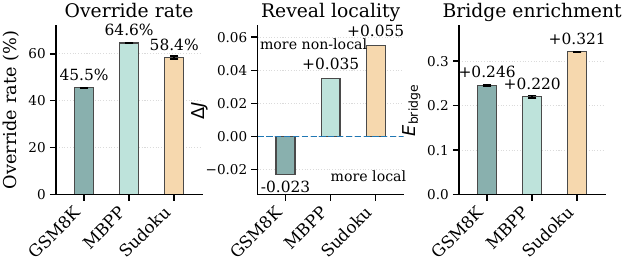}
    \vspace{-5mm}
    \caption{
    \textbf{Realized reveal behavior of CORD.}
    $\Delta J<0$ indicates more local decoding than confidence, whereas
    $\Delta J>0$ indicates more non-local decoding. Positive
    $E_{\mathrm{bridge}}$ indicates preferential selection of two-sided masked
    positions beyond their availability.
    }
    \label{fig:behavior_audit}
    \vspace{-4mm}
\end{wrapfigure}
As shown in Figure~\ref{fig:behavior_audit}, CORD substantially alters the confidence-based reveal order, overriding the
same-state confidence decision on $45.5\%$, $64.6\%$, and $58.4\%$ of decoding
steps on GSM8K, MBPP, and Sudoku, respectively. The resulting geometry is
task dependent: GSM8K becomes slightly more local
($\Delta J=-0.023$), whereas MBPP and Sudoku become more non-local
($+0.035$ and $+0.055$). All three tasks also exhibit positive bridge
enrichment, with $E_{\mathrm{bridge}}=+0.246$, $+0.220$, and $+0.321$,
respectively, showing that CORD actively favors two-sided infilling beyond
what would arise from candidate availability alone. Together, these results
show that the learned residuals induce substantial and task-specific changes
in the actual reveal trajectory.

\vspace{-2mm}
\subsection{Efficiency}
\label{sec:efficiency}
\vspace{-2mm}
We measure inference efficiency on a single NVIDIA H200 GPU using
LLaDA-8B-Instruct. Each example generates $256$ response tokens with a maximum
total sequence length of $1024$. We run each decoding method $20$ times and
report the mean and standard deviation of end-to-end generation latency.
Table~\ref{tab:efficiency} summarizes the results.
\begin{wraptable}{r}{0.47\linewidth}
\centering
\vspace{-6pt}
\caption{
\textbf{Inference efficiency on LLaDA-8B-Instruct.}
Latency (s) is measured on one H200 over $20$ runs with $256$ generated tokens
and a maximum sequence length of $1024$; we report mean $\pm$ standard deviation.
}
\vspace{-2mm}
\label{tab:efficiency}

\fontsize{7pt}{6pt}\selectfont
\setlength{\tabcolsep}{1pt}
\renewcommand{\arraystretch}{1.05}

\begin{tabular}{lcccc}
\toprule
Method
& Extra Fwd.
& Aux. Params
& Latency (s)
& Overhead \\
\midrule
Confidence
& 0 & 0 & 6.31 $\pm$ 0.31 & -- \\
Fixed residual
& 0 & 0 & 6.55 $\pm$ 0.28 & 3.80\% \\
\rowcolor{gray!20}
\textbf{CORD}
& 0 & 527,512 & 6.65 $\pm$ 0.30 & 5.39\% \\
\bottomrule
\end{tabular}

\vspace{-2mm}
\end{wraptable}
CORD introduces no additional backbone forward passes. Its
prompt-conditioned coefficient matrix is predicted once per response, while
direction, proximity, and bridge are computed from the current reveal mask at
each decoding step. As a result, CORD increases latency only modestly, from
$6.31$\,s for confidence decoding to $6.65$\,s, corresponding to a $5.39\%$
overhead. The fixed-residual control incurs $3.80\%$ overhead, indicating that
most of the additional cost comes from geometric score computation, while the
prompt-conditioned policy itself adds little inference cost.
The main additional cost of CORD occurs during training, where multiple
coefficient-space rollouts are evaluated for each prompt. At inference,
exploration is removed and a single coefficient matrix is reused throughout the
entire generation. 

\vspace{-2mm}
\section{Conclusion}
\vspace{-2mm}

We introduced \textsc{Cord}, a lightweight framework for adaptive reveal-order
selection in masked diffusion language models. Rather than learning token-level
orders from scratch, \textsc{Cord} retains confidence decoding as a stable
reference and learns prompt- and state-dependent geometric residuals over
direction, proximity, and two-sided infilling. Exploration occurs in a
low-dimensional coefficient space, enabling sequence-level optimization as a
contextual bandit while keeping the dLLM backbone frozen.
Across three masked diffusion models and five reasoning, code-generation, and
constraint-satisfaction benchmarks, \textsc{Cord} improves average task success
from $32.0\%$ to $48.2\%$ over confidence decoding and also outperforms
per-domain fixed residuals. Ablations confirm the importance of prompt
conditioning, state adaptation, and all three residual channels. Behavioral
audits further show substantial task-dependent changes in reveal locality and
two-sided infilling. Finally, \textsc{Cord} requires no additional backbone
forward passes and adds only $5.39\%$ inference latency. Overall, effective
diffusion decoding benefits from retaining confidence as a strong default while
learning \emph{when and how to depart from it}.

\subsection*{AI Use Statement}

Generative AI tools were used solely to assist with language editing, including
improving grammar, clarity, and presentation of author-written text. They were
not used to generate the research ideas, formulate the proposed method, design
the experiments, produce or analyze experimental results, or make scientific
claims. All AI-assisted edits were reviewed and revised by the authors to ensure
technical correctness, consistency with the conducted experiments, and
appropriate attribution of prior work. The authors take full responsibility for
the final content of the manuscript.

\subsection*{Ethics Statement}

This work studies decoding algorithms for masked diffusion language models and
does not involve human-subject experiments, collection of personally
identifiable information, or acquisition of new sensitive data. Our evaluation
uses established research benchmarks, and publicly available pretrained language models under their respective research-use terms. CORD changes the order in which masked positions are revealed but does not alter the underlying model parameters or introduce new model capabilities. Nevertheless, improvements in language-model decoding may inherit the broader risks associated with the underlying models, including generation of incorrect,
biased, or unsafe content.

\subsection*{Reproducibility Statement}

We provide the information required to reproduce the main experiments in the
paper and appendix. This includes the definitions and implementation details of
the confidence-anchored geometric residuals, the prompt-conditioned coefficient
policy, state-dependent coefficient modulation, coefficient-space exploration,
group-relative policy-gradient training, sequence-level reward construction,
and deterministic inference procedure. We additionally specify the evaluated
backbones, benchmarks and data splits, decoding budgets, training
hyperparameters, baseline configurations, task-specific verifiers, ablation
settings, and computational resources. All comparisons use frozen pretrained
backbones and follow the decoding and evaluation protocols described in
Appendix~\ref{app:setup}. The appendix further provides the mathematical
justification for the coefficient-space estimator and the behavioral analyses
used to characterize the learned reveal policies.
We have also submitted the complete implementation together with the
supplementary material, including the training, inference, evaluation, and
analysis code needed to reproduce the reported experiments.

\bibliography{iclr2027_conference}

@article{austin2021d3pm,
  title={Structured denoising diffusion models in discrete state-spaces},
  author={Austin, Jacob and Johnson, Daniel D and Ho, Jonathan and Tarlow, Daniel and Van Den Berg, Rianne},
  journal={Advances in neural information processing systems},
  volume={34},
  pages={17981--17993},
  year={2021}
}

@article{lou2024sedd,
  title={Discrete diffusion modeling by estimating the ratios of the data distribution},
  author={Lou, Aaron and Meng, Chenlin and Ermon, Stefano},
  journal={arXiv preprint arXiv:2310.16834},
  year={2023}
}

@article{sahoo2024mdlm,
  title={Simple and effective masked diffusion language models},
  author={Sahoo, Subham S and Arriola, Marianne and Schiff, Yair and Gokaslan, Aaron and Marroquin, Edgar and Chiu, Justin T and Rush, Alexander and Kuleshov, Volodymyr},
  journal={Advances in Neural Information Processing Systems},
  volume={37},
  pages={130136--130184},
  year={2024}
}

@article{shi2024md4,
  title={Simplified and generalized masked diffusion for discrete data},
  author={Shi, Jiaxin and Han, Kehang and Wang, Zhe and Doucet, Arnaud and Titsias, Michalis},
  journal={Advances in neural information processing systems},
  volume={37},
  pages={103131--103167},
  year={2024}
}

@inproceedings{ou2024radd,
  title={Your absorbing discrete diffusion secretly models the conditional distributions of clean data},
  author={Ou, Jingyang and Nie, Shen and Xue, Kaiwen and Zhu, Fengqi and Sun, Jiacheng and Li, Zhenguo and Li, Chongxuan},
  booktitle={International Conference on Learning Representations},
  volume={2025},
  pages={64972--65009},
  year={2025}
}

@inproceedings{nie2025llada,
  title={Large Language Diffusion Models},
  author={Nie, Shen and Zhu, Fengqi and You, Zebin and Zhang, Xiaolu and Ou, Jingyang and Hu, Jun and Zhou, Jun and Lin, Yankai and Wen, Ji-Rong and Li, Chongxuan},
  booktitle={Advances in Neural Information Processing Systems},
  volume={38},
  pages={50608--50646},
  year={2025}
}

@article{ye2025dream,
  title={{Dream 7B}: Diffusion Large Language Models},
  author={Ye, Jiacheng and Xie, Zhihui and Zheng, Lin and Gao, Jiahui and Wu, Zirui and Jiang, Xin and Li, Zhenguo and Kong, Lingpeng},
  journal={arXiv preprint arXiv:2508.15487},
  year={2025},
  url={https://arxiv.org/abs/2508.15487}
}

@article{survey2025dlm,
  title={A survey on diffusion language models},
  author={Li, Tianyi and Chen, Mingda and Guo, Bowei and Shen, Zhiqiang},
  journal={arXiv preprint arXiv:2508.10875},
  year={2025}
}

@article{kim2025trainworst,
  title={Train for the Worst, Plan for the Best: Understanding Token Ordering in Masked Diffusions},
  author={Kim, Jaeyeon and Shah, Kulin and Kontonis, Vasilis and Kakade, Sham and Chen, Sitan},
  journal={arXiv preprint arXiv:2502.06768},
  year={2025},
  url={https://arxiv.org/abs/2502.06768}
}

@article{peng2025p2,
  title={Path planning for masked diffusion model sampling},
  author={Peng, Fred Zhangzhi and Bezemek, Zachary and Patel, Sawan and Rector-Brooks, Jarrid and Yao, Sherwood and Bose, Avishek Joey and Tong, Alexander and Chatterjee, Pranam},
  journal={arXiv preprint arXiv:2502.03540},
  year={2025}
}

@inproceedings{liu2024ddpd,
  title={Think while you generate: Discrete diffusion with planned denoising},
  author={Liu, Sulin and Nam, Juno and Campbell, Andrew and St{\"a}rk, Hannes and Xu, Yilun and Jaakkola, Tommi and G{\'o}mez-Bombarelli, Rafael},
  booktitle={International Conference on Learning Representations},
  volume={2025},
  pages={81979--82026},
  year={2025}
}

@inproceedings{sahoo2025papl,
  title={Planner aware path learning in diffusion language models training},
  author={Peng, Zhangzhi and Bezemek, Zachary and Rector-Brooks, Jarrid and Zhang, Shuibai and Bronstein, Michael and Zhang, Anru and Bose, Joey and Tong, Alexander},
  booktitle={International Conference on Learning Representations},
  volume={2026},
  pages={126799--126844},
  year={2026}
}

@inproceedings{wu2025fastdllm,
  title={Fast-dllm: Training-free acceleration of diffusion llm by enabling kv cache and parallel decoding},
  author={Wu, Chengyue and Zhang, Hao and Xue, Shuchen and Liu, Zhijian and Diao, Shizhe and Zhu, Ligeng and Luo, Ping and Han, Song and Xie, Enze},
  booktitle={International Conference on Learning Representations},
  volume={2026},
  pages={57027--57051},
  year={2026}
}

@inproceedings{huang2025dcolt,
  title={Reinforcing the Diffusion Chain of Lateral Thought with Diffusion Language Models},
  author={Huang, Zemin and Chen, Zhiyang and Wang, Zijun and Li, Tiancheng and Qi, Guo-Jun},
  booktitle={Advances in Neural Information Processing Systems},
  volume={38},
  pages={152677--152710},
  year={2025}
}

@article{zhao2025diffpo,
  title={Diffpo: Training diffusion llms to reason fast and furious via reinforcement learning},
  author={Zhao, Hanyang and Liang, Dawen and Tang, Wenpin and Yao, David and Kallus, Nathan},
  journal={arXiv preprint arXiv:2510.02212},
  year={2025}
}

@article{jazbec2025unmask,
  title={Learning unmasking policies for diffusion language models},
  author={Jazbec, Metod and Olausson, Theo X and B{\'e}thune, Louis and Ablin, Pierre and Kirchhof, Michael and Monteiro, Jo{\~a}o and Turrisi, Victor and Ramapuram, Jason and Cuturi, Marco},
  journal={arXiv preprint arXiv:2512.09106},
  year={2025}
}

@article{du2025dultra,
  title={dultra: Ultra-fast diffusion language models via reinforcement learning},
  author={Chen, Shirui and Jiao, Jiantao and Ratliff, Lillian J and Zhu, Banghua},
  journal={arXiv preprint arXiv:2512.21446},
  year={2025}
}

@article{sas2026scheduling,
  title={Scheduling Thoughts: Learning the Order of Thought in Diffusion Language Models},
  author={Xu, Jiawei and Liu, Minghui and Agrawal, Aakriti and Chen, Yifan and Huang, Furong},
  journal={arXiv preprint arXiv:2606.23567},
  year={2026}
}

@inproceedings{zhu2025llada15,
  title={Llada 1.5: Variance-reduced preference optimization for large language diffusion models},
  author={Zhu, Fengqi and Wang, Rongzhen and Nie, Shen and Zhang, Xiaolu and Wu, Chunwei and Zhou, Jun and Lin, Yankai and Wen, Ji-Rong and Li, Chongxuan},
  booktitle={Proceedings of the 64th Annual Meeting of the Association for Computational Linguistics (Volume 1: Long Papers)},
  pages={11425--11460},
  year={2026}
}

@inproceedings{chang2022maskgit,
  title={Maskgit: Masked generative image transformer},
  author={Chang, Huiwen and Zhang, Han and Jiang, Lu and Liu, Ce and Freeman, William T},
  booktitle={2022 IEEE/CVF Conference on Computer Vision and Pattern Recognition (CVPR)},
  pages={11305--11315},
  year={2022},
  organization={IEEE}
}

@inproceedings{huang2026biases,
  title     = {Empirical Analysis of Decoding Biases in Masked Diffusion Models},
  author    = {Huang, Pengcheng and Liu, Tianming and Liu, Zhenghao and Yan, Yukun and Wang, Shuo and Xiao, Tong and Chen, Zulong and Sun, Maosong},
  booktitle = {Annual Meeting of the Association for Computational Linguistics},
  pages     = {6853--6876},
  year      = {2026},
  note      = {arXiv:2508.13021}
}

@article{ni2026flexibility,
  title={The Flexibility Trap: Why Arbitrary Order Limits Reasoning Potential in Diffusion Language Models},
  author={Ni, Zanlin and Wang, Shenzhi and Yue, Yang and Yu, Tianyu and Zhao, Weilin and Hua, Yeguo and Chen, Tianyi and Song, Jun and Yu, Cheng and Zheng, Bo and Huang, Gao},
  journal={arXiv preprint arXiv:2601.15165},
  year={2026},
  url={https://arxiv.org/abs/2601.15165v1}
}

@article{olausson2026temperatures,
  title={A Tale of Two Temperatures: Simple, Efficient, and Diverse Sampling from Diffusion Language Models},
  author={Olausson, Theo X and Jazbec, Metod and Wang, Xi and Solar-Lezama, Armando and Naesseth, Christian A and Mandt, Stephan and Nalisnick, Eric},
  journal={arXiv preprint arXiv:2604.09921},
  year={2026}
}

@article{asano2026where,
  title={Where-to-unmask: Ground-truth-guided unmasking order learning for masked diffusion language models},
  author={Asano, Hikaru and Kozuno, Tadashi and Saito, Kuniaki and Baba, Yukino},
  journal={arXiv preprint arXiv:2602.09501},
  year={2026}
}

@article{israel2025apd,
  title={Accelerating diffusion llms via adaptive parallel decoding},
  author={Israel, Daniel and Van den Broeck, Guy and Grover, Aditya},
  journal={Advances in neural information processing systems},
  volume={38},
  pages={52870--52888},
  year={2025}
}

@article{dsb2026,
  title={{DSB}: Dynamic Sliding Block Scheduling for Diffusion {LLMs}},
  author={Luo, Lizhuo and Li, Shenggui and Wen, Yonggang and Zhang, Tianwei},
  journal={arXiv preprint arXiv:2602.05992},
  year={2026},
  url={https://arxiv.org/abs/2602.05992}
}

@article{zhou2026attnsampler,
  title={Attention-based sampler for diffusion language models},
  author={Zhou, Yuyan and Hou, Kai Syun and Chen, Weiyu and Kwok, James},
  journal={arXiv preprint arXiv:2604.08564},
  year={2026}
}

@article{liu2026nisampling,
  title={{NI} Sampling: Accelerating Discrete Diffusion Sampling by Token Order Optimization},
  author={Liu, Enshu and Ning, Xuefei and Wang, Yu and Lin, Zinan},
  journal={arXiv preprint arXiv:2604.18471},
  year={2026},
  url={https://arxiv.org/abs/2604.18471}
}

@inproceedings{softmasked2026,
  title={Soft-Masked Diffusion Language Models},
  author={Hersche, Michael and Moor-Smith, Samuel and Hofmann, Thomas and Rahimi, Abbas},
  booktitle={International Conference on Learning Representations},
  volume={2026},
  pages={105685--105716},
  year={2026}
}

@article{adaptiveorder2026,
  title={Adaptive Order Policies for Masked Diffusion},
  author={Mohamud, Jama Hussein and Hasan, Mohsin and Ravanelli, Mirco and Bengio, Yoshua},
  journal={arXiv preprint arXiv:2606.00295},
  year={2026}
}

@inproceedings{benhamu2025entropy,
  title={Accelerated Sampling from Masked Diffusion Models via Entropy Bounded Unmasking},
  author={Ben-Hamu, Heli and Gat, Itai and Severo, Daniel and Nolte, Niklas S and Karrer, Brian},
  booktitle={Advances in Neural Information Processing Systems},
  volume={38},
  pages={55981--56007},
  year={2025}
}

@inproceedings{huang2026don,
  title={Don't Settle Too Early: Self-Reflective Remasking for Diffusion Language Models},
  author={Huang, Zemin and Wang, Yuhang and Chen, Zhiyang and Qi, Guo-Jun},
  booktitle={International Conference on Learning Representations},
  volume={2026},
  pages={56998--57026},
  year={2026}
}

@article{zhai2026core,
  title={Core: Context-robust remasking for diffusion language models},
  author={Zhai, Kevin and Mollah, Sabbir and Wang, Zhenyi and Shah, Mubarak},
  journal={arXiv preprint arXiv:2602.04096},
  year={2026}
}

@misc{yu2026revisedontfreezesamplermatched,
      title={Revise, Don't Freeze: Sampler-Matched Training for Self-Correcting Masked Diffusion Language Models}, 
      author={Longxuan Yu and Shaorong Zhang and Yu Fu and Hui Liu and Yue Dong and Greg Ver Steeg},
      year={2026},
      eprint={2606.01026},
      archivePrefix={arXiv},
      primaryClass={cs.CL},
      url={https://arxiv.org/abs/2606.01026}, 
}

@article{cobbe2021gsm8k,
  title={Training Verifiers to Solve Math Word Problems},
  author={Cobbe, Karl and Kosaraju, Vineet and Bavarian, Mohammad and Chen, Mark and Jun, Heewoo and Kaiser, Lukasz and Plappert, Matthias and Tworek, Jerry and Hilton, Jacob and Nakano, Reiichiro and Hesse, Christopher and Schulman, John},
  journal={arXiv preprint arXiv:2110.14168},
  year={2021}
}

@article{hendrycksmath2021,
  title={Measuring Mathematical Problem Solving With the MATH Dataset},
  author={Dan Hendrycks and Collin Burns and Saurav Kadavath and Akul Arora and Steven Basart and Eric Tang and Dawn Song and Jacob Steinhardt},
  journal={NeurIPS},
  year={2021}
}

@misc{chen2021evaluating,
      title={Evaluating Large Language Models Trained on Code},
      author={Mark Chen and Jerry Tworek and Heewoo Jun and Qiming Yuan and Henrique Ponde de Oliveira Pinto and Jared Kaplan and Harri Edwards and Yuri Burda and Nicholas Joseph and Greg Brockman and Alex Ray and Raul Puri and Gretchen Krueger and Michael Petrov and Heidy Khlaaf and Girish Sastry and Pamela Mishkin and Brooke Chan and Scott Gray and Nick Ryder and Mikhail Pavlov and Alethea Power and Lukasz Kaiser and Mohammad Bavarian and Clemens Winter and Philippe Tillet and Felipe Petroski Such and Dave Cummings and Matthias Plappert and Fotios Chantzis and Elizabeth Barnes and Ariel Herbert-Voss and William Hebgen Guss and Alex Nichol and Alex Paino and Nikolas Tezak and Jie Tang and Igor Babuschkin and Suchir Balaji and Shantanu Jain and William Saunders and Christopher Hesse and Andrew N. Carr and Jan Leike and Josh Achiam and Vedant Misra and Evan Morikawa and Alec Radford and Matthew Knight and Miles Brundage and Mira Murati and Katie Mayer and Peter Welinder and Bob McGrew and Dario Amodei and Sam McCandlish and Ilya Sutskever and Wojciech Zaremba},
      year={2021},
      eprint={2107.03374},
      archivePrefix={arXiv},
      primaryClass={cs.LG}
}

@article{austin2021program,
  title={Program Synthesis with Large Language Models},
  author={Austin, Jacob and Odena, Augustus and Nye, Maxwell and Bosma, Maarten and Michalewski, Henryk and Dohan, David and Jiang, Ellen and Cai, Carrie and Terry, Michael and Le, Quoc and others},
  journal={arXiv preprint arXiv:2108.07732},
  year={2021}
}

@misc{ye2025autoregressiondiscretediffusioncomplex,
      title={Beyond Autoregression: Discrete Diffusion for Complex Reasoning and Planning}, 
      author={Jiacheng Ye and Jiahui Gao and Shansan Gong and Lin Zheng and Xin Jiang and Zhenguo Li and Lingpeng Kong},
      year={2025},
      eprint={2410.14157},
      archivePrefix={arXiv},
      primaryClass={cs.CL},
      url={https://arxiv.org/abs/2410.14157}
}
\bibliographystyle{iclr2027_conference}

\newpage

\appendix

\section{Experimental Details}
\label{app:setup}

\subsection{Backbones}
\label{app:backbones}

We evaluate CORD on three masked diffusion language models:
LLaDA-8B-Instruct~\citep{nie2025llada},
LLaDA-1.5~\citep{zhu2025llada15}, and
Dream-7B~\citep{ye2025dream}. We train a separate CORD policy for each
backbone while keeping the underlying dLLM frozen.

\subsection{Benchmarks and Data Splits}
\label{app:data}

The training mixture is chosen to expose the policy to qualitatively different
response dependencies. Mathematical reasoning provides predominantly sequential
dependencies, code generation combines structural and local dependencies, and
Sudoku provides dense constraint interactions.

\begin{table}[!b]
\centering
\caption{Training and evaluation benchmarks.}
\label{tab:app_data}
\small
\renewcommand{\arraystretch}{1.08}
\begin{tabular}{c|c}
\textbf{config} & \textbf{value} \\
\midrule
math
& GSM8K, MATH \\
code
& HumanEval, MBPP \\
constraint satisfaction
& $4{\times}4$ Sudoku \\
training-domain sampling ratio
& $2{:}1{:}1$ (math : code : Sudoku) \\
math training prompts
& 4000 \\
code training prompts
& 500 \\
Sudoku training prompts
& 2000 \\
\end{tabular}
\end{table}

The stratified sampling ratio prevents the substantially smaller code collection
from becoming negligible in a mixed-domain update. Repeated visits to a training
prompt are not identical policy samples because each visit draws fresh
coefficient-space perturbations.

We do not truncate prompts that exceed the available context budget. Such
examples are removed before evaluation so that different reveal-order policies
operate on exactly the same input.

\subsection{Training Configuration}
\label{app:hparams}
For each prompt, CORD samples one coefficient-space action and keeps it fixed
for the complete rollout. Conditioned on this sample, reveal-position selection
is deterministic. The coefficient sample is detached before evaluating its
Gaussian log-density; otherwise the reparameterized sample and predicted mean
cancel in the score-function term, as discussed in
Appendix~\ref{app:estimator}.

Table~\ref{tab:app_train_setting} summarizes the training configuration.
CORD uses $B=3$ geometric residual channels: direction, proximity, and
bridge, and $J=8$ state basis functions. The coefficient-space action therefore
has dimension $n=BJ=24$. The state basis is as
defined in Sec.~\ref{sec:method}.

We train separate runs with total epoch budgets
$E\in\{20,30,40\}$ using AdamW with an initial learning rate of
$1\times10^{-3}$ and cosine decay. For CORD built on
LLaDA-8B-Instruct, training is performed on $8$ NVIDIA H200 GPUs and a
complete training run takes approximately $8$ hours for 40 epochs.

The final policy layer is zero-initialized, so the predictive mean initially
produces zero residual coefficients and CORD reduces exactly to confidence-based Top2-margin decoding. During training, Gaussian exploration is added only in the
low-dimensional coefficient space, with $\sigma$ annealed from $0.6$ to $0.15$
but never to zero. This exploration is necessary because top-$1$ reveal
decisions make the sequence-level reward piecewise constant with respect to the
coefficients; Appendix~\ref{app:estimator} details the corresponding
score-function estimator.

\begin{table}[t]
\centering
\caption{Training configuration for CORD.}
\label{tab:app_train_setting}
\small
\renewcommand{\arraystretch}{1.08}
\begin{tabular}{c|c}
\textbf{config} & \textbf{value} \\
\midrule
residual channels $B$
& $3$ \\
basis functions $J$
& $8$ \\
coefficient dimension $n=BJ$
& $24$ \\
policy hidden dimension
& $128$ \\
policy depth
& $2$ linear layers \\
activation
& SiLU \\
output-layer initialization
& zeros \\
group size $K$
& $8$ \\
prompts per update
& $64$ \\
training epochs $E$
& $\{20,30,40\}$ \\
optimizer
& AdamW \\
learning rate
& \{$1\times10^{-3}$, $5\times10^{-4}$, $1\times10^{-4}$\} \\
learning-rate schedule
& cosine decay \\
exploration scale $\sigma$
& $0.6\rightarrow0.15$ \\
exploration schedule
& linear decay over first $70\%$ of training \\
format reward $\lambda_{\mathrm{fmt}}$
& $0.1$ \\
answer-probability reward $\lambda_{\mathrm{ans}}$
& $0$ by default \\
\end{tabular}
\end{table}

\subsection{Decoding Configuration}
\label{app:decoding}

\begin{table}[t]
\centering
\caption{Default decoding configuration.}
\label{tab:app_decoding}
\small
\renewcommand{\arraystretch}{1.08}
\begin{tabular}{c|c}
\textbf{config} & \textbf{value} \\
\midrule
generation length $L$
& $512$ \\
number of decoding steps $T$
& $512$ \\
positions revealed per step
& $1$ \\
maximum sequence length
& $2048$ \\
decoding schedule
& pure any-order \\
position selection
& top-$1$ by reveal score \\
inference coefficient
& predictive mean $\boldsymbol{\mu}_\psi(x)$ \\

\end{tabular}
\end{table}

Prompts are formatted using the native chat template of each instruction-tuned
checkpoint. HumanEval inputs are passed in their standard function-signature and
docstring format. All experiments in the main paper use top-$1$ reveal: exactly
one masked response position is committed at each decoding step.

\subsection{Verifiers}
\label{app:verifier}

Sequence-level correctness is evaluated using task-specific deterministic
verifiers. The same verifier implementation is used during training and
evaluation.

\paragraph{GSM8K.}
We extract the committed final answer from the standard answer marker and
compare the normalized numerical value with the reference answer.

\paragraph{MATH.}
We extract the final boxed answer and compare it using exact matching,
normalization, and symbolic equivalence when parsing succeeds. Parse failures
receive zero correctness reward and are logged separately.

\paragraph{HumanEval and MBPP.}
Generated programs are executed against the corresponding test cases in an
isolated subprocess with a fixed timeout, memory limit, clean temporary
directory, and network access disabled. A completion is correct only if all
tests pass.

\paragraph{Sudoku.}
A generated board is accepted if it preserves all clues, assigns every cell,
and satisfies every row, column, and box constraint. We verify constraints
directly rather than comparing against one reference solution, allowing puzzles
with multiple valid solutions.

We separately record parse failures, execution timeouts, and verifier failures
to detect changes in measured reward caused by output-format or execution
pathologies rather than improved task performance.




\subsection{Comparison with Remasking-Based Decoders}
\label{sec:remask_comparison}

Recent work improves dLLM generation by allowing previously committed tokens
to be reopened and revised. Table~\ref{tab:remask_comparison} compares CORD
with representative methods from this line of work on overlapping reasoning
and code benchmarks. In contrast, CORD is strictly monotonic: once a token is
revealed, it is never remasked or revised, and the pretrained dLLM backbone
remains frozen.

Despite this restriction, CORD substantially outperforms several remasking
baselines in their reported results. Compared with CORE, CORD improves GSM8K,
MATH, HumanEval, and MBPP by $25.5\%$, $17.8\%$, $25.6\%$, and $24.1\%$,
respectively. CORE is training-free, but performs explicit inference-time
revision by probing previously committed tokens and introduces additional
revision forward passes beyond the standard decoding trajectory
\citep{zhai2026core}. CORD instead modifies only the ordering of still-masked
positions and requires no additional backbone evaluation.
CORD also exceeds SCOPE+D3IM on the
same four benchmarks, respectively. This comparison is notable because
SCOPE+D3IM explicitly equips the model for revision: D3IM allows visible tokens
to return to the masked state, while SCOPE performs sampler-matched
post-training so that the backbone learns under the self-correcting decoding
process. CORD obtains higher reported performance while leaving the pretrained
backbone unchanged and never reopening a committed token.
RemeDi represents a stronger form of learned self-correction. Its Remask-SFT
model with the learned remask sampler reports $86.3\%$, $52.2\%$, $71.3\%$,
and $57.8\%$ on GSM8K, MATH, HumanEval, and MBPP, respectively
\citep{huang2026don}. Unlike CORD, however, RemeDi explicitly trains the model
to identify and revise poor commitments through dedicated remasking
post-training, including Remask SFT, and augments generation with a learned
remask policy. CORD therefore does not match the strongest revision-trained
model, but reaches comparable performance without any
remasking capability or backbone update.

These results highlight two complementary ways to improve diffusion decoding.
Remasking methods attempt to \emph{repair} earlier commitments after they have
been made, often through additional revision computation or remask-aware
post-training. CORD instead improves the commitment itself by adapting
\emph{which still-masked position should be revealed next}. Its advantage over
CORE and SCOPE+D3IM suggests that better reveal-order control alone can recover
a substantial amount of performance that might otherwise require explicit
revision. 

\begin{table}[t]
\centering
\fontsize{8pt}{8pt}\selectfont
\setlength{\tabcolsep}{3pt}
\caption{
\textbf{Comparison with remasking-based decoding methods.}
Reported task performance (\%) on overlapping reasoning and code benchmarks.
CORD uses strictly monotonic decoding: once a token is revealed, it is never
remasked or revised. Results from prior work are reproduced from their
respective papers and are not directly comparable because model checkpoints,
prompting, and decoding budgets differ.
}
\label{tab:remask_comparison}

\renewcommand{\arraystretch}{1.08}

\begin{tabular}{lccrrrr}
\toprule
Method
& Remask
& Backbone Update
& GSM8K
& MATH-500
& HumanEval
& MBPP \\
\midrule

CORE~\citep{zhai2026core}
& \checkmark
& \xmark
& 52.7
& 17.1
& 17.1
& 24.8 \\

SCOPE+D3IM~\citep{yu2026revisedontfreezesamplermatched}
& \checkmark
& \checkmark
& 73.8
& 30.0
& 34.8
& 36.6 \\

RemeDi (Remask SFT)~\citep{huang2026don}
& \checkmark
& \checkmark
& 86.3
& 52.2
& 71.3
& 57.8 \\

\midrule
\rowcolor{gray!20}
\textbf{CORD}
& \xmark
& \xmark
& 78.2
& 34.9
& 42.7
& 48.9 \\
\bottomrule
\end{tabular}

\vspace{-1mm}
\end{table}

\section{Properties of Coefficient-Space Training}
\label{app:estimator}

We next justify the coefficient-space estimator used to train CORD. All
quantities are defined locally when introduced.

Fix a prompt $x$. The policy predicts a coefficient mean
$\boldsymbol{\mu}(x)\in\mathbb{R}^{n}$, where $n=BJ$, and samples
\begin{equation}
q_\sigma(\boldsymbol{\zeta}\mid x)
=
\mathcal{N}
\left(
\boldsymbol{\mu}(x),
\sigma^2I_n
\right).
\label{eq:app_q}
\end{equation}
Let $u$ denote all rollout randomness other than
$\boldsymbol{\zeta}$, such as token-sampling random numbers. We assume $u$ is
independent of the coefficient sample and of the policy parameters. For fixed
$x$ and $u$, let $r_x(\boldsymbol{\zeta},u)$ denote the sequence-level reward
obtained by the complete deterministic reveal trajectory induced by
$\boldsymbol{\zeta}$. We assume
$|r_x(\boldsymbol{\zeta},u)|\le R$ for some finite $R$ and that
$\boldsymbol{\mu}(x)$ is differentiable with respect to the policy parameters
$\psi$.

\subsection{Score-Function Gradient}

Define the Gaussian-smoothed training objective
\begin{equation}
J_\sigma(\psi)
=
\mathbb{E}_{x,u}
\mathbb{E}_{\boldsymbol{\zeta}
\sim q_\sigma(\cdot\mid x)}
\left[
r_x(\boldsymbol{\zeta},u)
\right].
\label{eq:app_objective}
\end{equation}

\begin{proposition}[Coefficient-space score-function gradient]
\label{prop:unbiased}
Under the bounded-reward assumption above,
\begin{equation}
\nabla_\psi J_\sigma
=
\mathbb{E}_{x,u,\boldsymbol{\zeta}}
\left[
r_x(\boldsymbol{\zeta},u)
\nabla_\psi
\log q_\sigma(\boldsymbol{\zeta}\mid x)
\right].
\label{eq:app_score}
\end{equation}
Moreover,
\begin{equation}
\nabla_{\boldsymbol{\mu}}
\log q_\sigma(\boldsymbol{\zeta}\mid x)
=
\frac{
\boldsymbol{\zeta}-\boldsymbol{\mu}(x)
}{
\sigma^2
}.
\label{eq:app_mu_score}
\end{equation}
\end{proposition}

\begin{proof}
For fixed $(x,u)$, write
\begin{equation}
F(\boldsymbol{\mu})
=
\int
r_x(\boldsymbol{\zeta},u)
q_\sigma(\boldsymbol{\zeta}\mid\boldsymbol{\mu})
\,d\boldsymbol{\zeta}.
\end{equation}
Because the reward is bounded and the derivative of a Gaussian location density
is a Gaussian multiplied by a linear factor, differentiation can be exchanged
with integration by dominated convergence on any compact neighborhood of
$\boldsymbol{\mu}$. Hence
\begin{equation}
\nabla_{\boldsymbol{\mu}}F
=
\int
r_x(\boldsymbol{\zeta},u)
q_\sigma(\boldsymbol{\zeta}\mid\boldsymbol{\mu})
\nabla_{\boldsymbol{\mu}}
\log q_\sigma(\boldsymbol{\zeta}\mid\boldsymbol{\mu})
\,d\boldsymbol{\zeta}.
\end{equation}
Averaging over $(x,u)$ and applying the chain rule through
$\boldsymbol{\mu}(x)=f_\psi(e(x))$ gives
Eq.~\ref{eq:app_score}.
\end{proof}

Importantly, this gradient differentiates the sampling density, not the
discrete reveal trajectory. No derivative of the verifier or discrete reveal
operation is required.

\subsection{Leave-One-Out Group Baseline}

For one prompt, CORD samples
$\boldsymbol{\zeta}^{(1)},\ldots,\boldsymbol{\zeta}^{(K)}$
independently from Eq.~\ref{eq:app_q}. All sibling rollouts share the same
non-coefficient randomness $u$. Let
\begin{equation}
r^{(i)}
=
r_x
\left(
\boldsymbol{\zeta}^{(i)},u
\right)
\end{equation}
and define
\begin{equation}
A^{(i)}
=
r^{(i)}
-
\frac{1}{K-1}
\sum_{j\ne i}
r^{(j)}.
\label{eq:app_loo}
\end{equation}

\begin{proposition}[Unbiased leave-one-out estimator]
\label{prop:loo}
The estimator
\begin{equation}
\widehat g_\psi
=
\frac{1}{K}
\sum_{i=1}^{K}
A^{(i)}
\nabla_\psi
\log
q_\sigma
\left(
\boldsymbol{\zeta}^{(i)}
\mid x
\right)
\label{eq:app_loo_estimator}
\end{equation}
is unbiased for the gradient in Eq.~\ref{eq:app_score}.
\end{proposition}

\begin{proof}
Condition on $(x,u)$. For rollout $i$, define
\begin{equation}
b^{(i)}
=
\frac{1}{K-1}
\sum_{j\ne i}r^{(j)}.
\end{equation}
Since $b^{(i)}$ depends only on coefficient samples other than
$\boldsymbol{\zeta}^{(i)}$, it is independent of
$\boldsymbol{\zeta}^{(i)}$ conditional on $(x,u)$. Therefore,
\begin{equation}
\mathbb{E}
\left[
b^{(i)}
\nabla_\psi
\log q_\sigma
\left(
\boldsymbol{\zeta}^{(i)}
\mid x
\right)
\middle|
x,u
\right]
=
0,
\end{equation}
because the score of a normalized density has zero expectation. The reward term
is unbiased by Proposition~\ref{prop:unbiased}. Averaging over the group and
then over $(x,u)$ completes the proof.
\end{proof}

Sharing $u$ therefore does not invalidate the leave-one-out baseline. The
coefficient samples remain conditionally independent given the shared
randomness.

\subsection{Equal-Reward Groups}

If every sampled rollout in a group receives the same reward, then
Eq.~\ref{eq:app_loo} gives
\begin{equation}
A^{(i)}=0
\qquad
\text{for every }i,
\end{equation}
and the group contributes exactly zero gradient.

It is useful to contrast this with an unbaselined score estimator in coefficient
space. Suppose, for fixed $(x,u)$, the reward is constant as a function of the
coefficient action,
\begin{equation}
r_x(\boldsymbol{\zeta},u)\equiv c.
\end{equation}
For the gradient with respect to the mean
$\boldsymbol{\mu}$, the unbaselined estimator is
\begin{equation}
\widehat g_{\mu,0}
=
\frac{c}{K\sigma}
\sum_{i=1}^{K}
\boldsymbol{\epsilon}^{(i)},
\qquad
\boldsymbol{\epsilon}^{(i)}
\sim
\mathcal{N}(\mathbf{0},I_n),
\end{equation}
so
\begin{equation}
\mathbb{E}
\left[
\widehat g_{\mu,0}
\right]
=
0,
\qquad
\mathbb{E}
\left[
\left\|
\widehat g_{\mu,0}
\right\|_2^2
\right]
=
\frac{c^2n}{K\sigma^2}.
\label{eq:app_constant_reward_noise}
\end{equation}

Equation~\ref{eq:app_constant_reward_noise} applies to the
\emph{coefficient-mean gradient}; the gradient with respect to network
parameters additionally contains the Jacobian
$\nabla_\psi\boldsymbol{\mu}(x)$. The leave-one-out baseline removes this
pure score noise exactly whenever the observed group rewards are identical.

\subsection{Why We Do Not Standardize the Group Advantage}

Consider binary rewards and suppose that $m$ of the $K$ sibling rollouts are
successful. With
\begin{equation}
\widehat p=\frac{m}{K},
\qquad
\widehat s
=
\sqrt{
\widehat p(1-\widehat p)
},
\end{equation}
the magnitude of the standardized leave-one-out advantage is
\begin{equation}
\frac{
\left|
A^{(i)}
\right|
}{
\widehat s
}
=
\begin{cases}
\dfrac{K}{K-1}
\sqrt{
\dfrac{K-m}{m}
},
&
r^{(i)}=1,
\\[10pt]
\dfrac{K}{K-1}
\sqrt{
\dfrac{m}{K-m}
},
&
r^{(i)}=0.
\end{cases}
\label{eq:app_standardized_adv}
\end{equation}

Thus, a lone minority outcome in an otherwise nearly unanimous group receives
a weight of order $\sqrt{K}$, while a balanced group receives weights of order
one. At unanimity, $\widehat s=0$.

More importantly, the scale $\widehat s$ depends on
$r^{(i)}$ and hence on the action
$\boldsymbol{\zeta}^{(i)}$. The conditional independence used in the proof of
Proposition~\ref{prop:loo} therefore no longer applies after dividing by the
within-group standard deviation. Such standardization is not guaranteed to
preserve the unbiased score-function estimator, so we leave the advantages
unstandardized.

\subsection{Piecewise-Constant Reward Landscape}

The top-1 reveal operation partitions coefficient space into regions that
induce the same reveal trajectory.

\begin{proposition}[Local constancy]
\label{prop:plateau}
Fix $(x,u)$ and a coefficient vector
$\boldsymbol{\zeta}_0$. Suppose that at every decoding step induced by
$\boldsymbol{\zeta}_0$, the highest and second-highest reveal scores are
distinct. Then there exists an open neighborhood
$\mathcal{U}$ of $\boldsymbol{\zeta}_0$ such that every
$\boldsymbol{\zeta}\in\mathcal{U}$ induces the same reveal trajectory,
generated response, and sequence reward.
\end{proposition}

\begin{proof}
At the first decoding step, the state is fixed. The sampled coefficient vector
is reshaped linearly into $C_{\boldsymbol{\zeta}}$, and
Eq.~\ref{eq:residual_coefficients} maps it continuously through $\tanh$ to
the residual coefficients. Since the normalized confidence and geometric
scores are fixed for the current state, each reveal score
$S(p,s;\boldsymbol{\zeta})$ is a continuous function of
$\boldsymbol{\zeta}$.

A strict gap between the largest and second-largest reveal scores therefore
persists in some open neighborhood of $\boldsymbol{\zeta}_0$, so the same
position is selected throughout that neighborhood. Given the same selected
position and the same rollout randomness $u$, the revealed token and next
decoding state are identical. Repeating this argument over the finite decoding
trajectory and intersecting the resulting neighborhoods gives the desired open
set $\mathcal{U}$.
\end{proof}

\begin{corollary}[Zero pathwise gradient inside a decoding region]
\label{cor:pathwise}
At every coefficient vector satisfying the conditions of
Proposition~\ref{prop:plateau},
\begin{equation}
\nabla_{\boldsymbol{\zeta}}
r_x(\boldsymbol{\zeta},u)
=
0.
\end{equation}
\end{corollary}

This explains why a pathwise derivative through
$\boldsymbol{\zeta}
=
\boldsymbol{\mu}(x)
+
\sigma\boldsymbol{\epsilon}$
cannot provide the desired learning signal: away from reveal-order boundaries,
the terminal reward is locally constant even though moving probability mass
between decoding regions can change the expected reward.

\paragraph{Detaching the sampled action.}
The same observation has an implementation consequence. If
$\boldsymbol{\zeta}
=
\boldsymbol{\mu}
+
\sigma\boldsymbol{\epsilon}$
is left attached to $\boldsymbol{\mu}$ while evaluating its Gaussian score,
then
\begin{equation}
\boldsymbol{\zeta}
-
\boldsymbol{\mu}
=
\sigma\boldsymbol{\epsilon}
\end{equation}
is algebraically independent of $\boldsymbol{\mu}$ and the intended
score-function gradient is lost. We therefore stop gradients through the sampled
coefficient action before evaluating its log-density.

\subsection{Exploration and Vanishing Learning Signal}

Proposition~\ref{prop:plateau} also characterizes the limit
$\sigma\rightarrow0$.

\begin{proposition}[Consistency of the smoothed objective]
\label{prop:sigma_consistency}
Fix $(x,u)$ and suppose the policy mean
$\boldsymbol{\mu}(x)$ lies in a locally constant decoding region. Then
\begin{equation}
\lim_{\sigma\rightarrow0}
\mathbb{E}_{\boldsymbol{\zeta}
\sim
\mathcal{N}
(
\boldsymbol{\mu}(x),
\sigma^2I
)}
\left[
r_x(\boldsymbol{\zeta},u)
\right]
=
r_x
\left(
\boldsymbol{\mu}(x),u
\right).
\end{equation}
\end{proposition}

\begin{proof}
By Proposition~\ref{prop:plateau}, there exists $\rho>0$ such that the reward is
constant inside
$B(\boldsymbol{\mu},\rho)$. Since the reward is bounded,
\begin{equation}
\left|
\mathbb{E}[r_x]
-
r_x(\boldsymbol{\mu},u)
\right|
\le
2R
\Pr
\left[
\boldsymbol{\zeta}
\notin
B(\boldsymbol{\mu},\rho)
\right],
\end{equation}
and the probability on the right tends to zero as
$\sigma\rightarrow0$.
\end{proof}

\begin{proposition}[Vanishing exploration removes the group signal]
\label{prop:sigma_collapse}
Under the same conditions, for any fixed $K$,
\begin{equation}
\Pr
\left[
r^{(1)}
=
\cdots
=
r^{(K)}
\right]
\rightarrow
1
\qquad
\text{as }
\sigma\rightarrow0.
\end{equation}
Moreover, the second moment of the leave-one-out
coefficient-mean estimator tends to zero.
\end{proposition}

\begin{proof}
All $K$ samples fall inside the locally constant ball with probability tending
to one, which proves the first claim. On this event every advantage is zero.
Outside the ball, the advantages are bounded, while Gaussian tail probabilities
decay exponentially in $1/\sigma^2$ and dominate the polynomial
$1/\sigma^2$ factor in the Gaussian score. Hence the second moment of the
coefficient-mean estimator also tends to zero.
\end{proof}

This result does not imply that exploration should never decrease. It shows only
that annealing $\sigma$ to zero while learning is still required eventually
eliminates transitions between distinct decoding regions.

\subsection{Common Random Numbers}

When token generation is stochastic, we use the same token-sampling randomness
$u$ for all coefficient actions associated with one prompt.

For fixed $u$, define
\begin{equation}
m(u)
=
\mathbb{E}_{\boldsymbol{\zeta}}
\left[
r_x(\boldsymbol{\zeta},u)
\right],
\qquad
v(u)
=
\operatorname{Var}_{\boldsymbol{\zeta}}
\left[
r_x(\boldsymbol{\zeta},u)
\right].
\end{equation}

\begin{proposition}[Variance reduction from shared randomness]
\label{prop:crn}
For two independently sampled coefficient actions,
\begin{equation}
\begin{aligned}
&
\operatorname{Var}
\left[
r_x
\left(
\boldsymbol{\zeta}^{(i)},u^{(i)}
\right)
-
r_x
\left(
\boldsymbol{\zeta}^{(j)},u^{(j)}
\right)
\right]
\\
&\quad -
\operatorname{Var}
\left[
r_x
\left(
\boldsymbol{\zeta}^{(i)},u
\right)
-
r_x
\left(
\boldsymbol{\zeta}^{(j)},u
\right)
\right]
\\
&=
2
\operatorname{Var}_u
\left[
m(u)
\right]
\ge
0.
\end{aligned}
\label{eq:app_crn}
\end{equation}
\end{proposition}

\begin{proof}
Write
\begin{equation}
r_x(\boldsymbol{\zeta},u)
=
m(u)
+
e(\boldsymbol{\zeta},u),
\qquad
\mathbb{E}_{\boldsymbol{\zeta}}
\left[
e(\boldsymbol{\zeta},u)
\mid u
\right]
=
0.
\end{equation}
With shared $u$, the two $m(u)$ terms cancel, yielding variance
$2\mathbb{E}_u[v(u)]$. With independent
$u^{(i)}$ and $u^{(j)}$, the difference additionally contains
$m(u^{(i)})-m(u^{(j)})$, whose variance is
$2\operatorname{Var}_u[m(u)]$. This yields
Eq.~\ref{eq:app_crn}.
\end{proof}

Under greedy token generation, $u$ is degenerate and the two schemes coincide.

\section{Why Confidence Alone Does Not Determine Reveal Value}
\label{app:confidence}

CORD uses confidence as its base decoder, but augments it with geometric
residuals. This section gives a simple motivation for why current per-position
confidence need not identify the position whose revelation is most useful for
future decoding. Because CORD reveals one token at a time, the issue is not a
simultaneous product-of-marginals approximation. Rather, the choice of which
position to reveal changes the context on which all subsequent predictions are
conditioned.

Let $Q_p(\cdot\mid s)$ denote the current predictive marginal at masked position
$p$. A confidence-only reveal rule ranks positions using a statistic
$c(Q_p)$ of this current marginal, such as maximum probability, top-two margin,
or entropy. Such a statistic describes how decisive the model is at $p$, but
does not by itself describe how revealing $p$ will affect the predictions at
other positions.

To make this distinction explicit, consider a reference conditional
distribution over the remaining masked variables and define the downstream
influence of revealing position $p$ by
\begin{equation}
I(p;s)
=
\mathbb{E}_{Y_p\sim P_p(\cdot\mid s)}
\left[
\sum_{q\in\mathcal{M}(s)\setminus\{p\}}
\mathrm{KL}
\left(
P_q(\cdot\mid s,Y_p)
\middle\|
P_q(\cdot\mid s)
\right)
\right].
\label{eq:reveal_influence}
\end{equation}
This quantity is only a conceptual device: CORD does not estimate it during
decoding. It captures the fact that two positions with similar current
confidence can differ substantially in how much revealing them changes the
future conditioning context.

\begin{proposition}[Marginal confidence does not determine reveal influence]
\label{prop:confidence_influence}
There exist two joint distributions over two binary variables with identical
current marginals, and therefore identical values for every confidence statistic
computed only from those marginals, but different reveal influence
$I(1;s)$.
\end{proposition}

\begin{proof}
Let $P^{(0)}$ be the independent uniform distribution over
$\{0,1\}^2$ and let $P^{(1)}$ be uniform over
$\{(0,0),(1,1)\}$. Both have
\begin{equation}
P(Y_1=0)=P(Y_1=1)
=
P(Y_2=0)=P(Y_2=1)
=
\frac12,
\end{equation}
so every statistic computed solely from the two current marginals is identical
under $P^{(0)}$ and $P^{(1)}$.

Under $P^{(0)}$, observing $Y_1$ leaves the distribution of $Y_2$ unchanged, so
\begin{equation}
I^{(0)}(1;s)=0.
\end{equation}
Under $P^{(1)}$, observing $Y_1$ determines $Y_2$. Hence
\begin{equation}
I^{(1)}(1;s)
=
\log 2
\end{equation}
when the KL divergence uses natural logarithms. Thus identical current
marginals can correspond to different downstream effects of revealing the same
position.
\end{proof}

The proposition does not imply that geometric information recovers the unknown
joint distribution, nor does CORD attempt to estimate
Eq.~\ref{eq:reveal_influence}. It only shows that a reveal rule based solely
on current marginal confidence cannot, in general, determine the downstream
effect of a reveal decision.

\subsection{What CORD Changes}

CORD keeps the frozen dLLM's confidence score as the base ranking and adds
residual biases derived from response geometry. Direction encodes a sequential
prior, proximity distinguishes local continuation from non-local seeding, and
the bridge signal identifies positions already surrounded by revealed context.
These signals are not intended as estimators of latent dependence. Instead,
they provide structured information unavailable to a confidence statistic that
depends only on the current marginal at each position.

With one-token-at-a-time decoding, the role of order is straightforward. If
position $p_t$ is revealed at state $s_t$, the next model evaluation is
conditioned on the updated state
\begin{equation}
s_{t+1}
=
s_t
\cup
\{
(p_t,\widehat y_{p_t})
\}.
\end{equation}
Choosing a different $p_t$ can therefore produce a different sequence of future
marginals even when the current confidence ranking is similar. CORD modifies
this conditioning path through the reveal order; it does not alter the frozen
token predictor itself.

\subsection{Reveal-Impact Diagnostic}
\label{app:reveal_impact}

For offline analysis, the one-token-at-a-time setting permits a direct diagnostic
of how strongly each reveal changes the model's remaining predictions. Let
$p_t$ be the position revealed at step $t$ and let $s_{t+1}$ be the resulting
state. We define
\begin{equation}
g_t
=
\frac{1}{
|\mathcal{M}(s_{t+1})|
}
\sum_{q\in\mathcal{M}(s_{t+1})}
D
\left(
\pi_q(\cdot\mid s_{t+1}),
\pi_q(\cdot\mid s_t)
\right),
\label{eq:reveal_impact}
\end{equation}
where $D$ is a distributional discrepancy such as Jensen--Shannon divergence.
Because top-1 decoding already performs a new backbone evaluation after each
reveal, both distributions are available from consecutive decoding states.

The trajectory-average impact is
\begin{equation}
\overline g
=
\frac{1}{T-1}
\sum_{t=1}^{T-1}
g_t.
\label{eq:mean_reveal_impact}
\end{equation}
This quantity is not a training objective and is not claimed to estimate the
true joint dependence structure. It is an internal sensitivity diagnostic: a
large $g_t$ means that revealing $p_t$ substantially changes the model's
predictions at positions that remain masked. Comparing this quantity across
confidence decoding and CORD helps characterize whether learned geometric
departures tend to select positions with different downstream conditioning
effects.

\section{Visualizations}
\label{app:viz}

\begin{figure}[H]
\centering
\includegraphics[trim={160 190 200 180}, clip, width=0.93\linewidth]{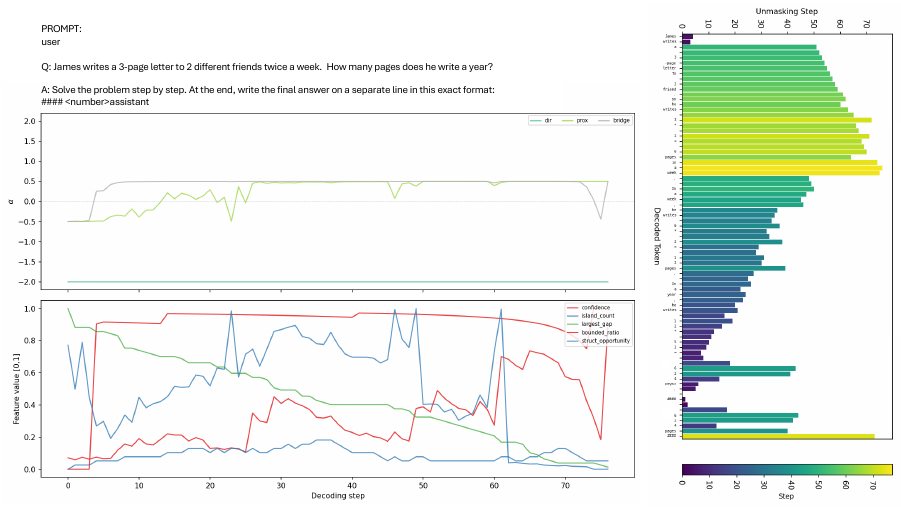}
\caption{
\textbf{A GSM8K rollout.}
\emph{Top:} token-unmasking timeline for a held-out mathematical reasoning
example. The realized reveal order is predominantly sequential, but with a few
targeted jumps that expose useful later positions before the entire prefix has
been completed. \emph{Bottom:} the corresponding coefficient and state-feature
trajectories. The direction coefficient remains strongly negative throughout the
rollout, indicating a persistent left-to-right bias, while the bridge
coefficient stays positive and the proximity coefficient gradually increases.
Together, these signals produce a mostly local reasoning trajectory with
occasional non-local or two-sided reveals when the evolving context makes them
useful.
}
\label{fig:revealmap}
\end{figure}

\begin{figure}[H]
\centering
\includegraphics[trim={160 190 200 180}, clip, width=0.93\linewidth]{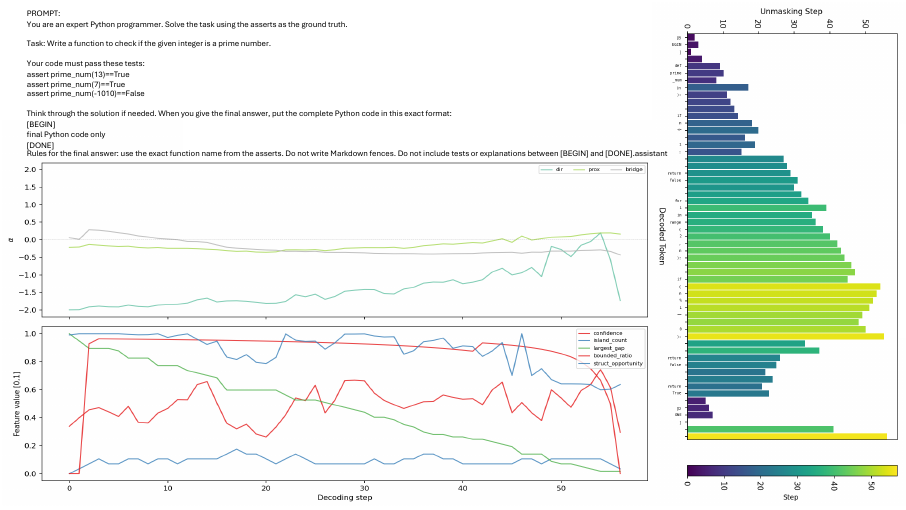}
\caption{
\textbf{A MBPP rollout.}
\emph{Top:} token-unmasking timeline for a held-out code-generation example.
The reveal order is again largely structured and directional, but includes a
small number of non-local reveals that help establish later code structure.
\emph{Bottom:} the corresponding coefficient and state-feature trajectories.
Here the direction coefficient provides the dominant bias, while the proximity
and bridge coefficients remain weaker and evolve more smoothly. The resulting
trajectory reflects a coding regime in which the model largely builds the
program in order, while still allowing selective departures from strict
sequential decoding when they improve the surrounding context.
}
\label{fig:diag}
\end{figure}

We provide qualitative rollout visualizations to illustrate how CORD translates
its learned residual policy into concrete reveal orders. Each figure shows one
representative held-out example. The upper panel is a token-unmasking timeline:
each decoded token is paired with the decoding step at which it is revealed,
so the plot directly exposes the realized reveal order. The lower panel shows
the corresponding coefficient trajectories
$\alpha_{\mathrm{dir}}$, $\alpha_{\mathrm{prox}}$, and
$\alpha_{\mathrm{bridge}}$, together with the normalized state features that
drive them. These examples complement the quantitative behavioral audit in
Sec.~\ref{sec:behavior} by showing how the learned residuals evolve within a
single rollout.

\section{Algorithmic Description}
\label{app:training_algorithm}

CORD separates stochastic exploration during training from deterministic
decoding at inference. For each training prompt, the policy predicts the mean
of a Gaussian distribution in the $24$-dimensional coefficient space. We draw
$K$ coefficient samples from this distribution, and each sample is held fixed
for an entire decoding rollout. Although the sampled coefficient matrix is
fixed within a rollout, the resulting residual coefficients remain
state-dependent because they are evaluated against the evolving state features
$f(s_t)$ at every decoding step. Sequence-level rewards are then used to update
only the lightweight CORD policy; the dLLM backbone remains frozen.

\begin{algorithm}[H]
\SetAlgoLined
\KwIn{frozen dLLM $\theta$; CORD policy $f_\psi$; group size $K$;
exploration scale $\sigma$}

\For{each training epoch}{
    sample stratified prompt batches using the $2{:}1{:}1$ domain ratio\;

    \ForEach{prompt $x$}{
        obtain prompt representation $e(x)$ from the frozen backbone\;

        predict coefficient mean
        $\boldsymbol{\mu}_\psi(x)
        \leftarrow
        f_\psi(e(x))
        \in\mathbb{R}^{24}$\;

        \For{$i=1,\ldots,K$}{
            sample
            $\boldsymbol{\epsilon}^{(i)}
            \sim\mathcal{N}(\mathbf{0},I)$\;

            set
            \[
            \boldsymbol{\zeta}^{(i)}
            \leftarrow
            \operatorname{detach}\!\left(
            \boldsymbol{\mu}_\psi(x)
            +
            \sigma\boldsymbol{\epsilon}^{(i)}
            \right)
            \]
            \;

            reshape
            $\boldsymbol{\zeta}^{(i)}$
            into
            $C^{(i)}\in\mathbb{R}^{3\times 8}$\;

            initialize the response block as fully masked\;

            \For{$t=1,\ldots,T$}{
                run the frozen dLLM once to obtain token predictions\;

                compute the current state features $f(s_t)$\;

                compute
                $\boldsymbol{\alpha}^{(i)}_t
                =
                \boldsymbol{\alpha}_{\max}
                \odot
                \tanh(C^{(i)}f(s_t))$\;

                compute the confidence-anchored residual score
                $S(p,s_t)$ for all masked positions\;

                reveal
                $p_t^\star=\arg\max_{p\in\mathcal{M}(s_t)}S(p,s_t)$\;
            }

            compute the sequence-level reward $r^{(i)}$\;
        }

        compute leave-one-out group-relative advantages
        $A^{(1)},\ldots,A^{(K)}$\;
    }

    update $\psi$ using the coefficient-space policy-gradient objective\;
    keep $\theta$ frozen\;
}

\KwOut{trained policy parameters $\psi$}
\caption{\textsc{Cord} training in coefficient space.}
\label{alg:cord_train}
\end{algorithm}

At inference, exploration is removed entirely. Given a prompt $x$, CORD
evaluates the policy once and uses its predictive mean,
\begin{equation}
C(x)
=
\operatorname{reshape}_{3\times 8}
\left(\boldsymbol{\mu}_\psi(x)\right),
\end{equation}
for the entire response. Importantly, fixing $C(x)$ does not make the reveal
strategy static. At decoding step $t$, the evolving partial response produces
a new state representation $f(s_t)$ and hence new residual coefficients,
\begin{equation}
\boldsymbol{\alpha}(x,s_t)
=
\boldsymbol{\alpha}_{\max}
\odot
\tanh\!\left(C(x)f(s_t)\right).
\end{equation}
CORD then adds the corresponding direction, proximity, and bridge residuals to
the confidence score and reveals the highest-scoring masked position. Thus, the
coefficient matrix is predicted only once per response, while the reveal
geometry adapts throughout decoding without stochastic sampling or additional
backbone evaluations.

\begin{algorithm}[H]
\SetAlgoLined
\KwIn{prompt $x$; frozen dLLM $\theta$; trained CORD policy $f_\psi$}

obtain prompt representation $e(x)$\;

compute
$\boldsymbol{\mu}_\psi(x)=f_\psi(e(x))$\;

set
$C(x)=
\operatorname{reshape}_{3\times8}
(\boldsymbol{\mu}_\psi(x))$\;

initialize the response block as fully masked\;

\For{$t=1,\ldots,T$}{
    run the frozen dLLM once to obtain token predictions\;

    compute state features $f(s_t)$\;

    compute
    $\boldsymbol{\alpha}(x,s_t)
    =
    \boldsymbol{\alpha}_{\max}
    \odot
    \tanh(C(x)f(s_t))$\;

    compute $S(p,s_t)$ for all
    $p\in\mathcal{M}(s_t)$\;

    reveal
    $p_t^\star
    =
    \arg\max_{p\in\mathcal{M}(s_t)}
    S(p,s_t)$\;
}

\KwOut{completed response}
\caption{\textsc{Cord} inference.}
\label{alg:cord_inference}
\end{algorithm}

\end{document}